%% file: im2cad.tex
\documentclass[10pt,twocolumn,letterpaper]{article}

\usepackage{cvpr}
\usepackage{times}
\usepackage{epsfig}
\usepackage{graphicx}
\usepackage{amsmath}
\usepackage{amssymb}
\usepackage{float}
\usepackage{booktabs}
\usepackage{xcolor,colortbl}
\usepackage{subfig}

\usepackage[pagebackref=true,breaklinks=true,letterpaper=true,colorlinks,bookmarks=false]{hyperref}

\cvprfinalcopy 

\def\cvprPaperID{xxxx} 

\begin{document}

\title{IM2CAD}

\author{
Hamid Izadinia \\
University of Washington
\and
Qi Shan \\
Zillow Group
\and
Steven M. Seitz \\
University of Washington
}

\teaser{
\vspace{1 mm}
\noindent\makebox[.48\textwidth][c]{
\minipage[c]{0.48\textwidth}
  \includegraphics[width=\linewidth]{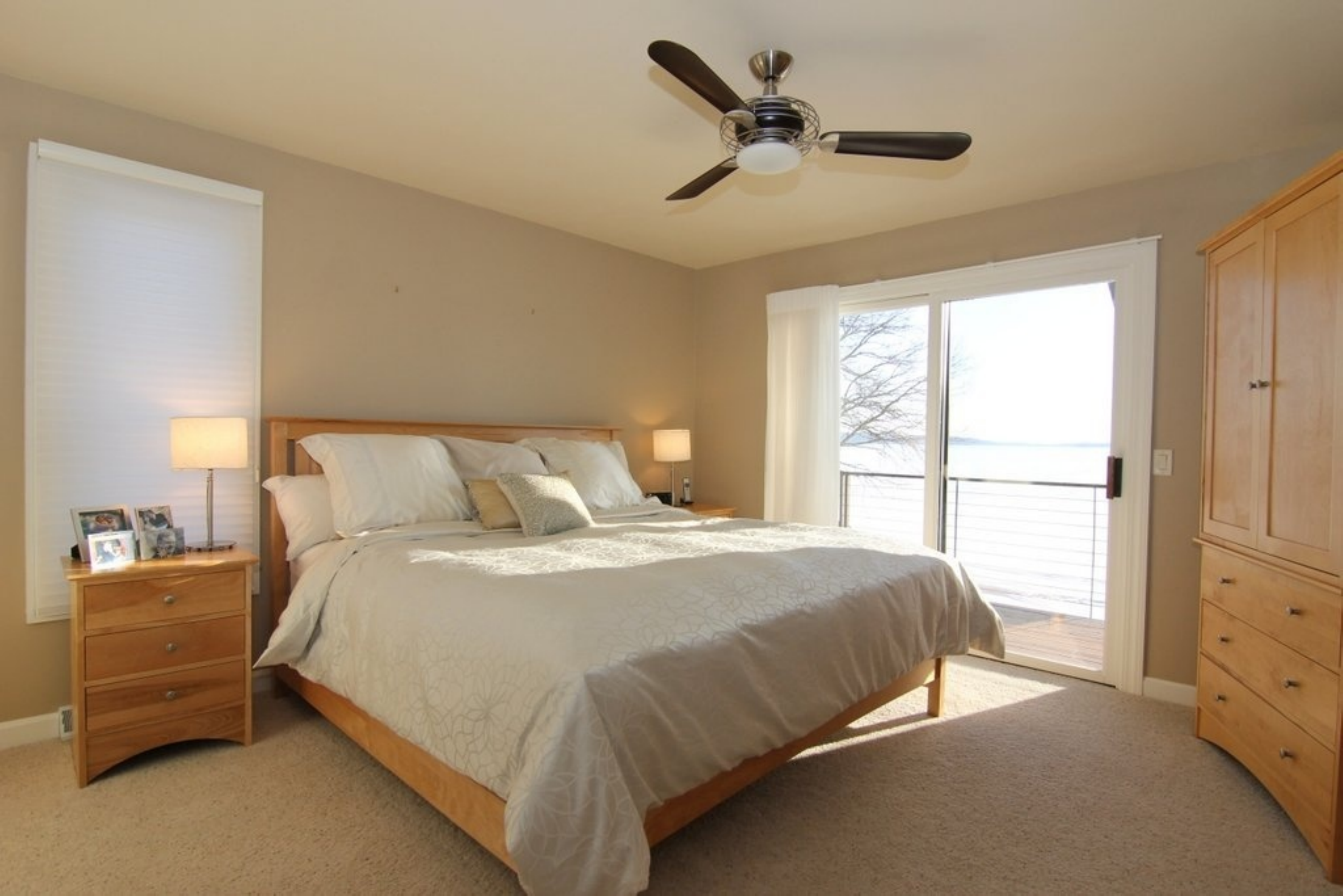}
\endminipage
}
\noindent\makebox[.48\textwidth][c]{
\minipage{0.48\textwidth}
  \includegraphics[width=\linewidth]{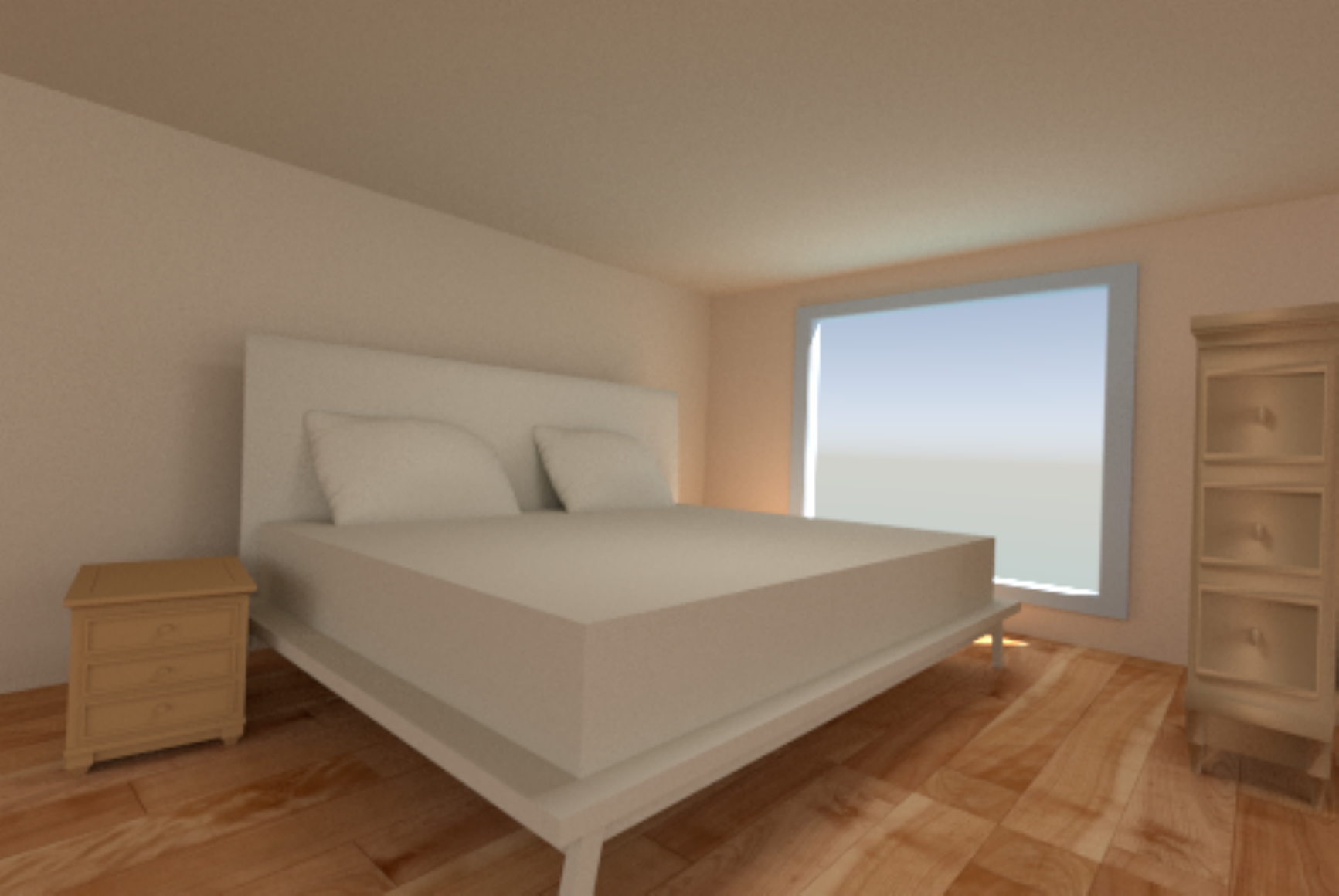}
\endminipage
}
\hfill

  \vspace{-3 mm}
   \caption{\normalsize IM2CAD takes a single photo of a real scene (left), and automatically reconstructs its 3D CAD model (right).}
   \label{fig:teaser}
\vspace{-1 mm}
}

\maketitle
\thispagestyle{empty}
\pagestyle{empty}

\begin{abstract}
Given a single photo of a room and a large database of furniture CAD models, our goal is to reconstruct a scene that is as similar as possible to the scene depicted in the photograph, and composed of objects drawn from the database.  
We present a completely automatic system to address this {\em IM2CAD} problem that produces high quality results on challenging imagery from interior home design and remodeling websites. Our approach iteratively optimizes the placement and scale of objects in the room to best match scene renderings to the input photo, using image comparison metrics trained via deep convolutional neural nets.  By operating jointly on the full scene at once, we account for inter-object occlusions. We also show the applicability of our method in standard scene understanding benchmarks where we obtain significant improvement. 
\end{abstract}

\vspace{-5 mm}

\input{intro}

\input{related}

\input{method}

\input{result}

\input{conclusion}

\section*{Acknowledgements}

This work was supported by funding from National Science Foundation grant IIS-1250793, Google, and the UW Animation Research Labs.

{\small
\bibliographystyle{ieee}
\bibliography{im2cad}
}

\end{document}

%% file: intro.tex
\section{Introduction}
\label{sec:intro}

\begin{figure}
  \centering
  \begin{tabular}{ccc}
  \subfloat[]{\includegraphics[width=.13\textwidth]{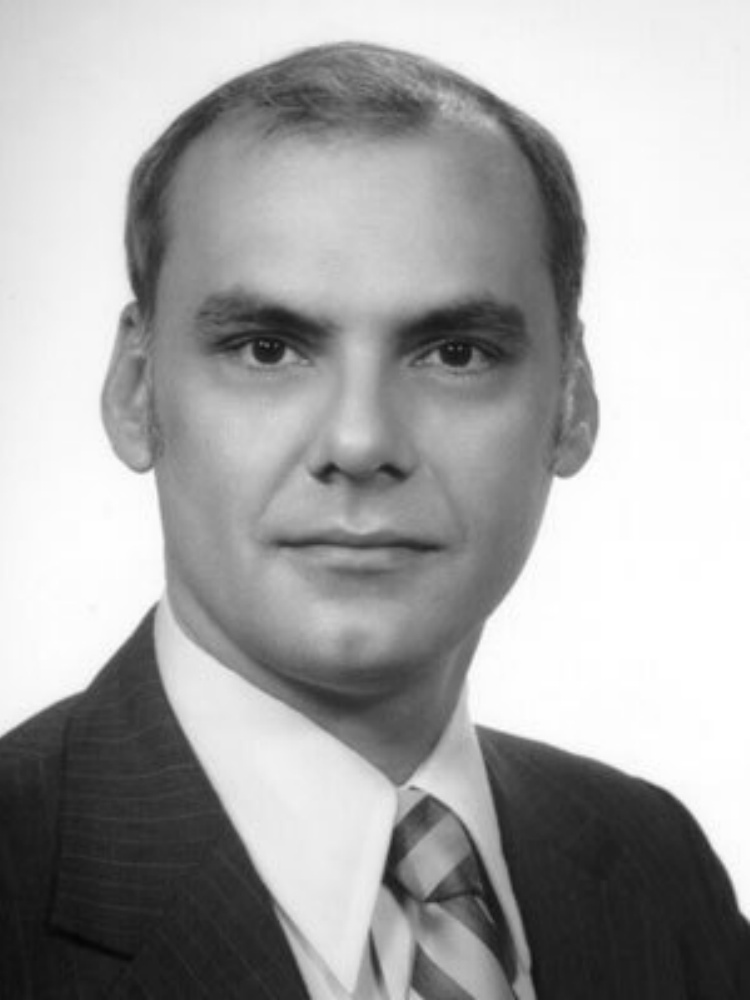}} &
  \subfloat[]{\includegraphics[width=.14\textwidth]{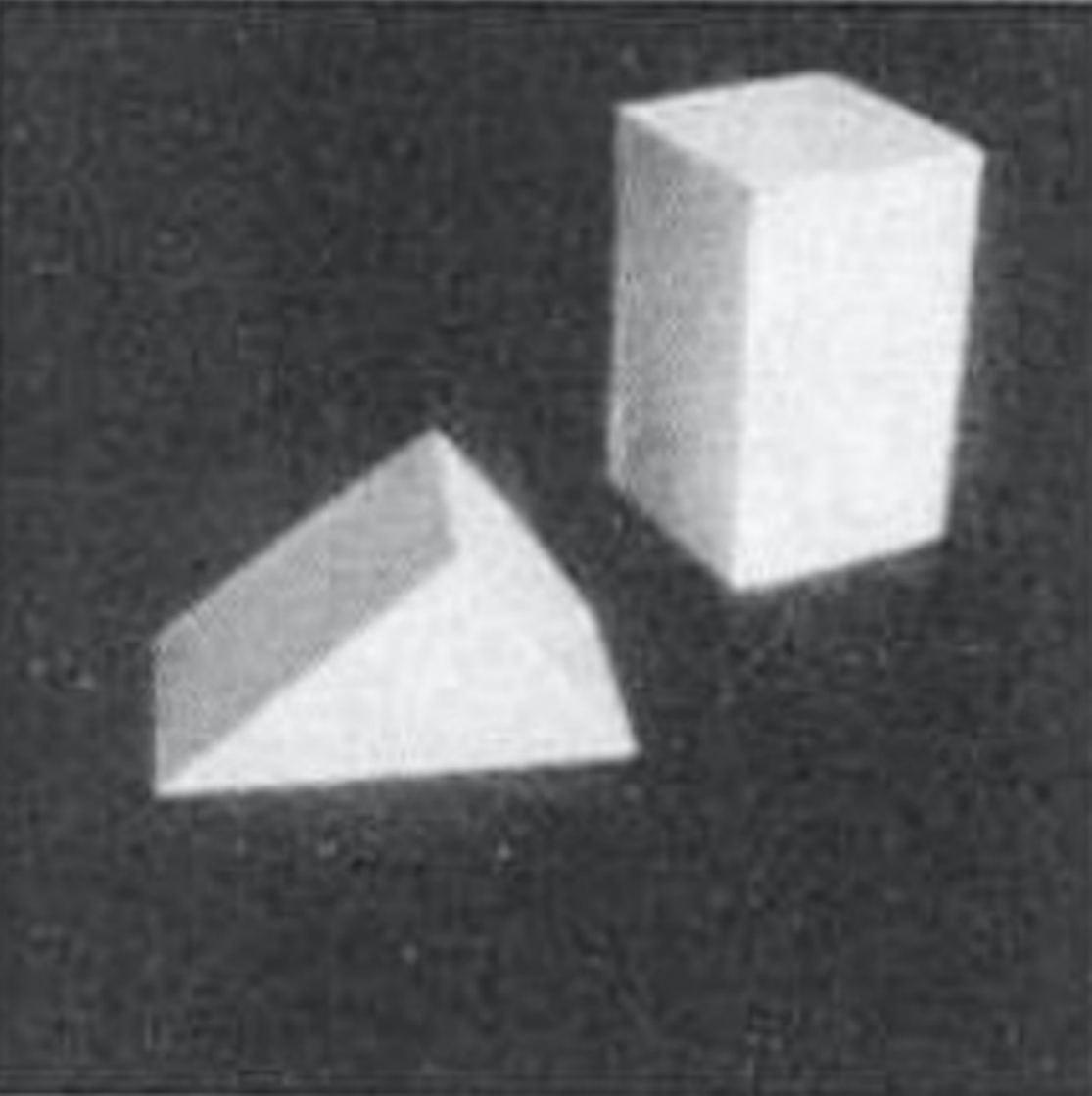}} &
  \subfloat[]{\includegraphics[width=.14\textwidth]{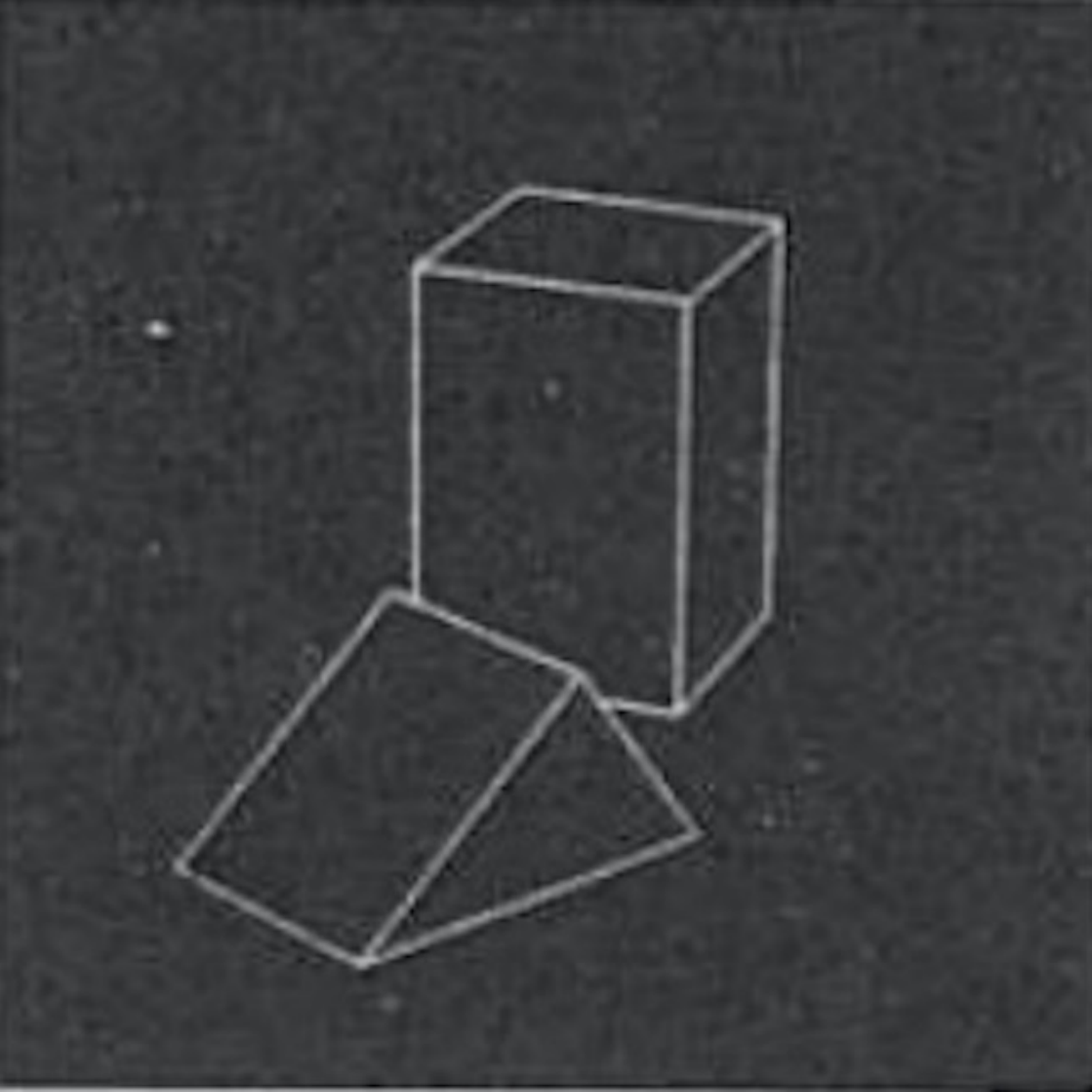}} \\
  \end{tabular}
  \vspace{-4.1 mm}
  \caption{Lawrence Roberts's (a) 1963 system took an input photo (b) and computed a 3D scene, rendered to a novel viewpoint (c). }
\label{fig:roberts}
\vspace{-6 mm}
\end{figure}

In his 1963 Ph.D. thesis, Lawrence Roberts \cite{roberts1963thesis} demonstrated a system that infers a 3D scene from a single photo (Figure~\ref{fig:roberts}).  Leveraging a database of known 3D objects, his system analyzed edges in the image to infer the locations and orientations of these objects in the scene.  Unlike the vast majority of modern 3D reconstruction techniques, which capture only visible surfaces, Robert's method was capable of inferring back-facing and occluded surfaces, object segments, and recognized which objects are present.

\begin{figure*}[t]
	\centering
	\includegraphics[clip, trim=0cm 0cm 0cm 0cm,width=.9\textwidth]{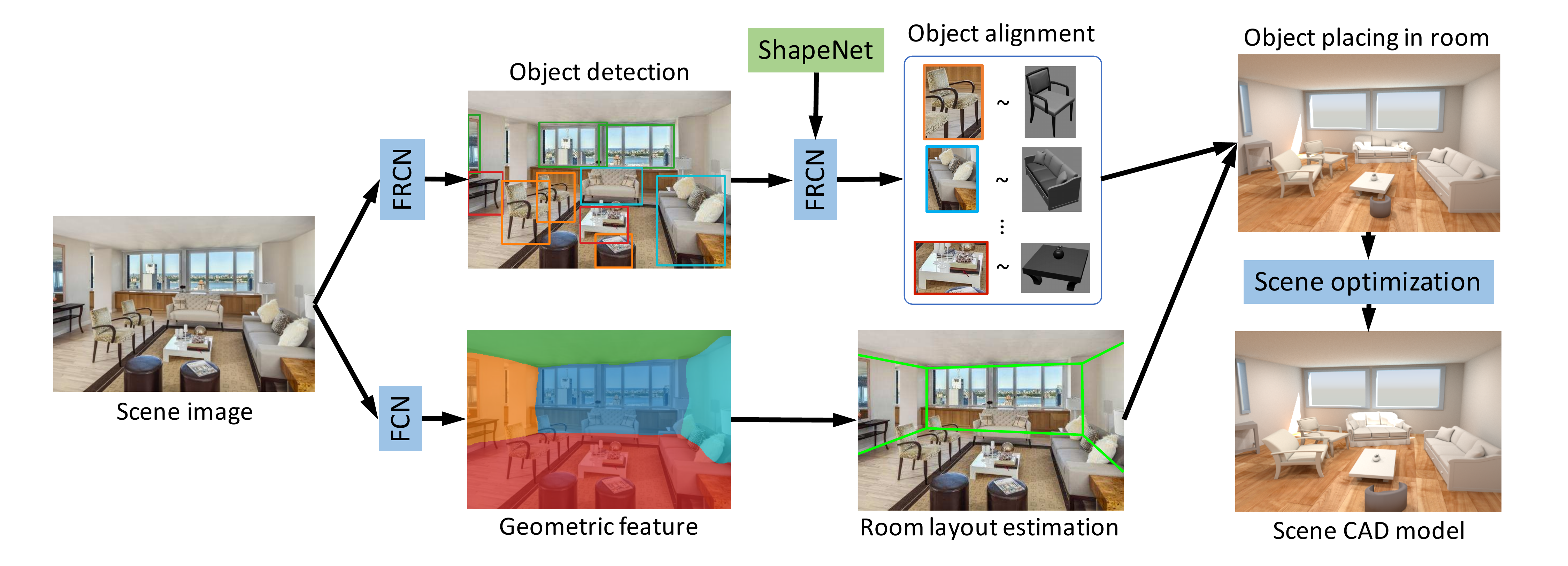}
	\vspace{-5 mm}
	\caption{System overview:  an input image (left) is processed through a series of steps to produce a scene CAD model (bottom right).}
	\label{fig:overview}
	\vspace{-6 mm}
\end{figure*}

While Robert's method was visionary, more than
a half century of subsequent research in computer vision has still not yet led to practical extensions of his approach that work reliably on realistic images and scenes.  One major limitation is the need for an accurate, a priori 3D model of each object in the scene.  While a chair model, e.g., is not hard to come by, obtaining exact 3D models of every chair in the world is not presently feasible.
A further challenge is the need to reliably match between features in photographs and CAD models, particularly when the model does not exactly match the object photographed.

We therefore introduce a variant of Robert's original problem, that we call {\em IM2CAD}, 
in which the goal is to reconstruct a scene that is {\bf as similar as possible} to the scene depicted in a photograph, 
where the reconstruction is composed of objects drawn from a database of available 3D object models.  
For example, the bed in Fig.~\ref{fig:teaser} resembles but does not exactly match the one in the input photograph at left, as we did not have that specific bed in the database. While our results are not perfect, they represent a significant step forward to achieving Robert's vision on real-world imagery.
Producing CAD models of real scenes has applications for virtual reality (VR), augmented reality (AR), robotics, games, and education.

Our work builds on a number of recent advances in the computer vision and graphics research community.  First, we leverage ShapeNet \cite{shapenet}, which contains millions of 3D models of objects, including thousands of different chairs, tables, and other household items.  This dataset is a game-changer for 3D scene understanding research, and was key to enabling our work.  Second, we use state-of-the-art object recognition algorithms \cite{ren2015faster} to identify common objects like chairs, tables, windows, etc.;  these methods work impressively well in practice.  Third, we leverage deep features trained by convolutional neural nets (CNNs)~\cite{NIPS2012_4824} to reliably match between photographs and CAD renderings \cite{aubry2014seeing,kholgade20143d,salas2013slam++,Huang2015}.  Finally, we build on recent research on room reconstruction \cite{Hedau09,dclee2009geometric,mallya15}.

Our main contribution is a fully automatic system that produces full-scene CAD models (room + furniture) from a single photo.  While many of the technical ingredients of our system draw heavily from prior work (as detailed in the previous paragraph), we also contribute noteworthy technical advances on room modeling and scene optimization.  Our room modeling approach produces significant improvement on standard benchmarks.
Our novel full-scene optimization approach iteratively adjusts the placement and scale of objects to best align rendered photos with input images, operating jointly on the full scene at once, and accounting for inter-object occlusions.  
Our models include semantics (e.g. ``table'', ``chair'') segmented into objects, and take only a few bytes to represent, encoded as a collection of ShapeNet object IDs and transformations that define position, orientation and scale. We evaluate our performance on scene understanding using the datasets of~\cite{Hedau09}, LSUN~\cite{lsun}, SUN RGB-D~\cite{song2015sun} and 3DGP~\cite{choi2015indoor}. We show significant improvements in the 2D and 3D room layout estimation as well as 3D object location using only single RGB images.

%% file: related.tex
\section{Related Work}
\label{sec:related}

The last decade has seen renewed interest in single-image 3D modeling, following the work of Hoiem et al.,~\cite{hoiem05a} and Saxena et al.,~\cite{saxena2005}. Single-image modeling of indoor scenes has enjoyed significant recent progress, with a series of papers on room-shape estimation (floor, walls, ceiling), e.g., \cite{Hedau09,dclee2009geometric,mallya15,Dasgupta_2016_CVPR,ren2016coarse} that yield increasingly good results. Our approach for room shape estimation obtains competitive results.
More recently, researchers have moved beyond walls, and toward approximating furniture in the room using {\em cuboids} \cite{xiao2012localizing,panocontext2014,choi2015indoor,gupta2010estimating,pero2012bayesian,Schwing_2013_ICCV}. While the cuboid based approach avoids the need for object databases, the resulting models are primitive and do not accurately depict scene appearance.

Another closely related line of research is 3D object and pose recognition of chairs and other objects \cite{aubry2014seeing,kholgade20143d,salas2013slam++,lim2014fpm,Huang2015,tulsiani2015viewpoints,bansal2016marr,wu2016single}.  These methods can produce very accurate alignment of a single object to a photograph or depth image. Our work leverages similar 3D object recognition techniques, combined with room shape estimation, to jointly solve for all of the objects in the room in a way that accounts for inter-object occlusions. Our work also builds upon recent advancements of research on object detection from single images~\cite{girshick2014rich,ren2015faster}.  

Researchers have explored a variety of techniques to automatically compute CAD scene models using non-photographic means, e.g., using example based approaches~\cite{fisher2012scenesynth}, utilizing text descriptions \cite{chang2014scenegen}, and optimizing for furniture arrangements in a given space~\cite{yu2011makeithome,merrell2011interactive}. These approaches rely on analyzing location and pose correlations between furniture types, based on analyzing databases of scene models.  Collecting such data is a challenge, and therefore these approaches can greatly benefit from our solution which generates more comprehensive and plausible indoor models in a fully automatic fashion.

The closest works to ours are~\cite{satkin20153dnn} and~\cite{liu2015model} which find the best matching 3D scene model to a given image. Our system is a significant advance in a number of ways. In particular, ~\cite{satkin20153dnn} requires a complete scene in the database that matches each image. Hence, their approach can be thought of as ``3D scene retrieval,''  whereas we reconstruct each scene from scratch, using a database of furniture (not scene) models. The latter allows for a much broader range of reconstructable scenes. While~\cite{liu2015model} reconstructs the scene by placing individual pieces of furniture, they make a number of limiting assumptions (axis aligned furniture, no walls, easy-to-segment objects), operate on a much smaller database (180 models), and do not demonstrate as broad a range of results. Both of~\cite{satkin20153dnn} and~\cite{liu2015model} use hand-crafted features, while our proposed method uses CNN feature which is learned end-to-end on just image data. Guo et al.~\cite{guo2015predicting} render a synthesized model of the scene using RGBD (depth) images while our method only uses RGB information. The synthesized rooms produced by~\cite{guo2015predicting} have low fidelity in terms of object details while we retrieve the detailed ShapeNet CAD model for each object.

In the context of 3D prediction, several previous approaches estimate the depth and surface normals of visible surfaces from a single image~\cite{Eigen_2015_ICCV, zeisl2014discriminatively, bansal2016marr}. In contrast, our approach does not require dense surface normal estimation but is capable of estimating both visible and invisible surfaces through joint estimation of room and object CAD models.

%% file: method.tex

\begin{figure}[t]
	\captionsetup[subfigure]{labelformat=empty,farskip=-13pt,position=bottom}
	\centering
	\subfloat[]{\includegraphics[width=.12\textwidth, height=.06\textheight]{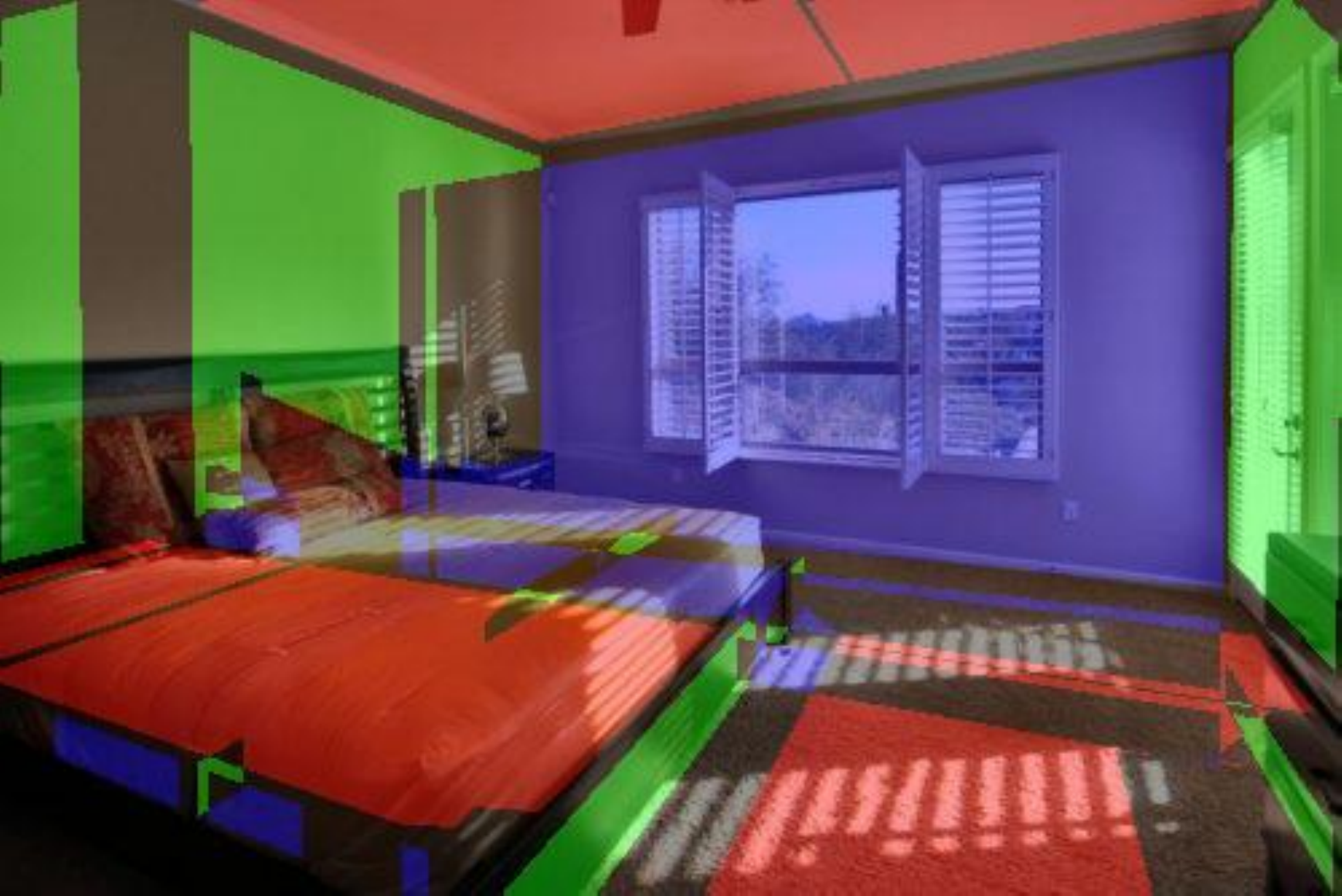}}
	\subfloat[]{\includegraphics[width=.12\textwidth, height=.06\textheight]{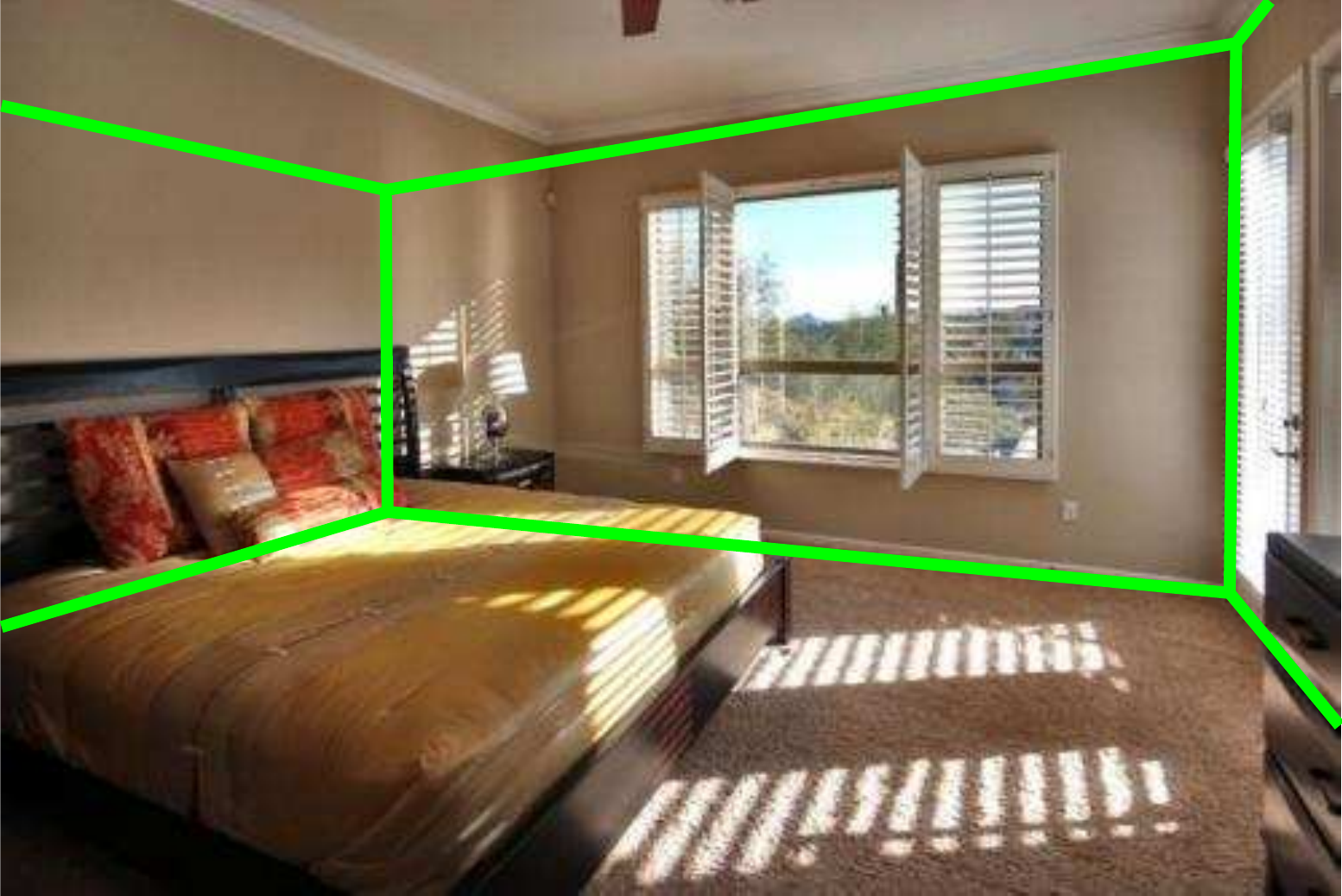}}
	\subfloat[]{\includegraphics[width=.12\textwidth, height=.06\textheight]{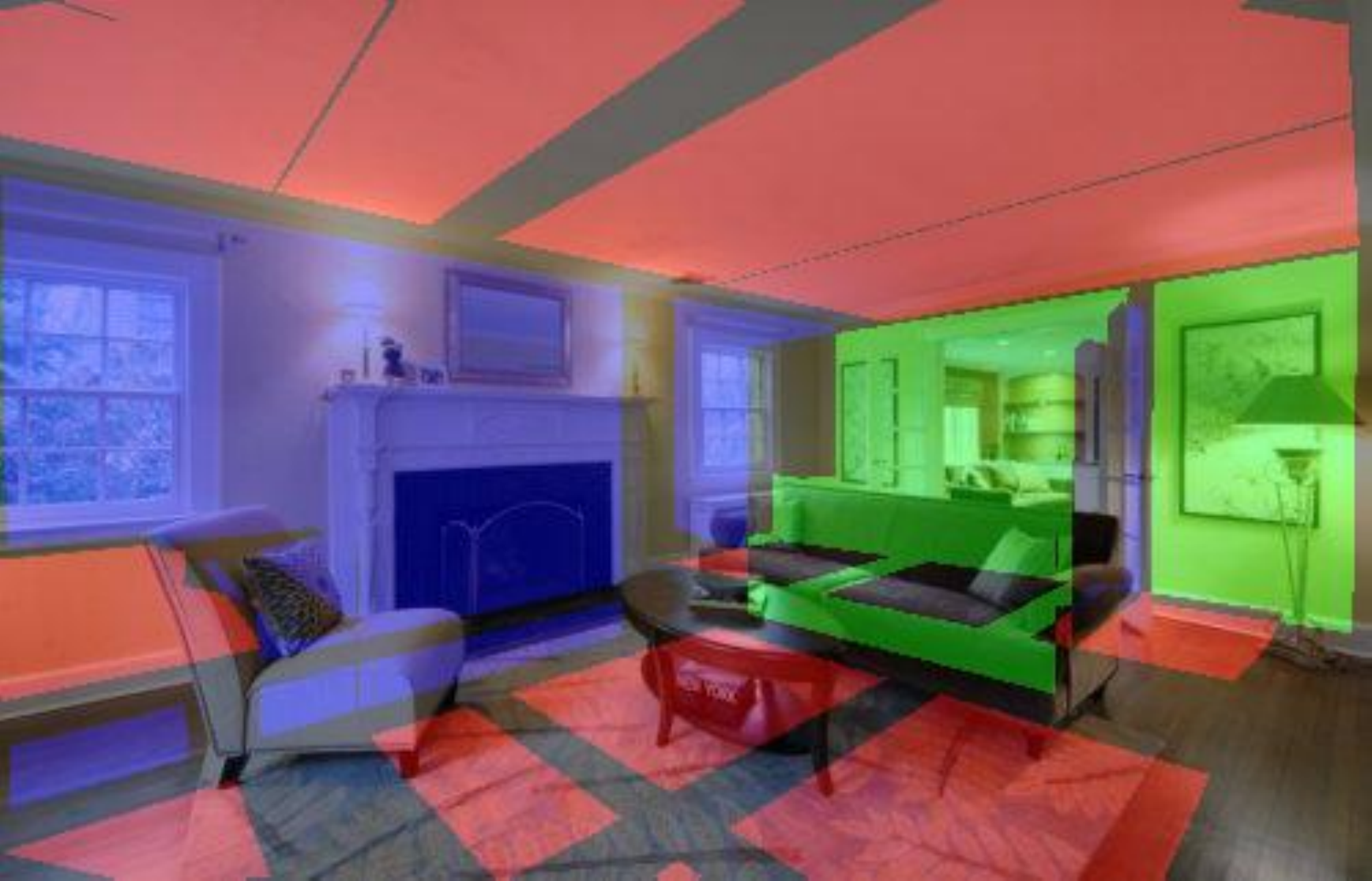}}
	\subfloat[]{\includegraphics[width=.12\textwidth, height=.06\textheight]{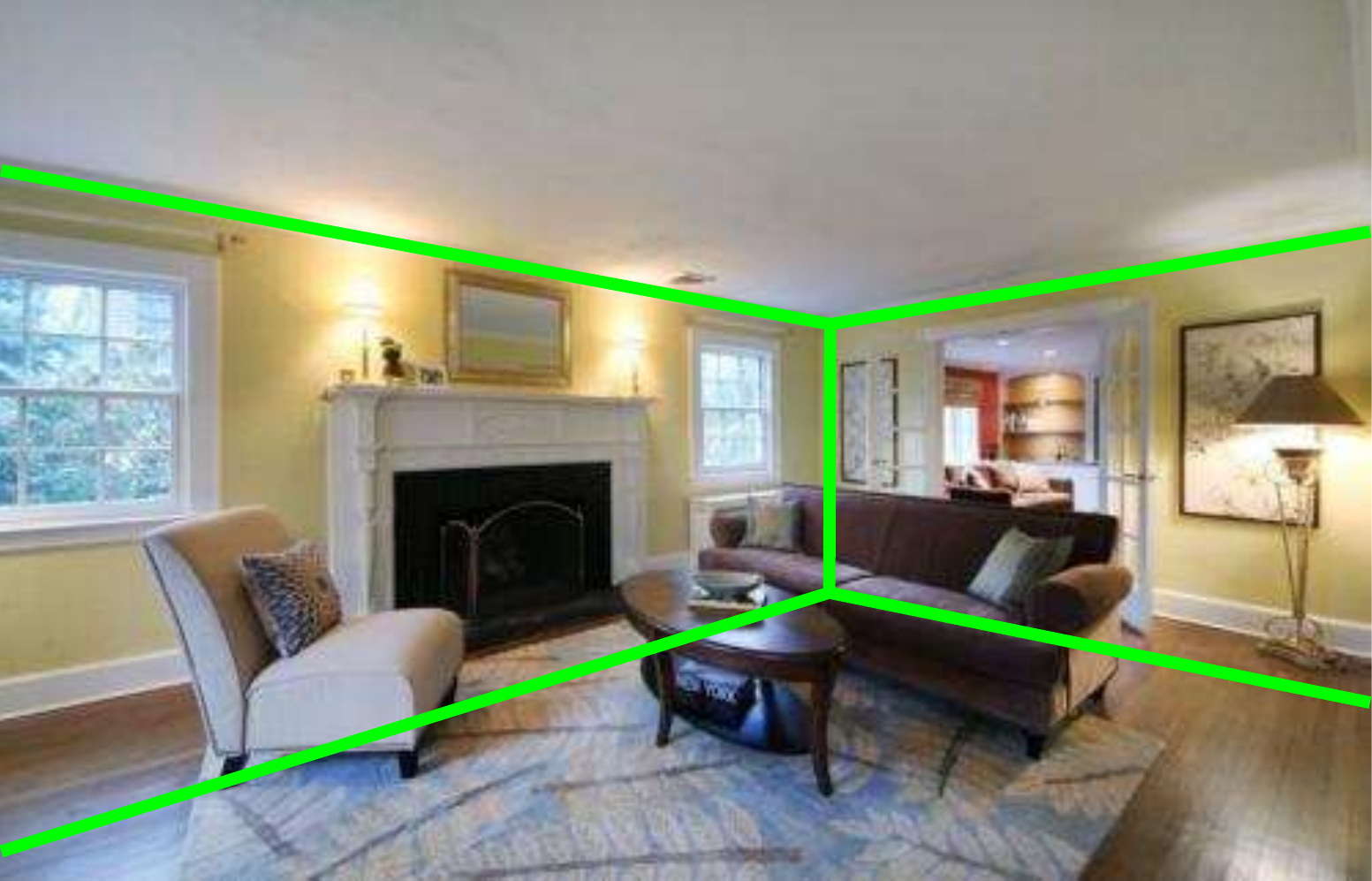}}\quad
	\subfloat[]{\includegraphics[clip, trim=2cm 9cm 2cm 9cm,width=.5\textwidth]{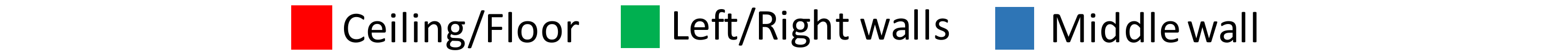}}\quad
	\subfloat[]{\includegraphics[width=.12\textwidth, height=.06\textheight]{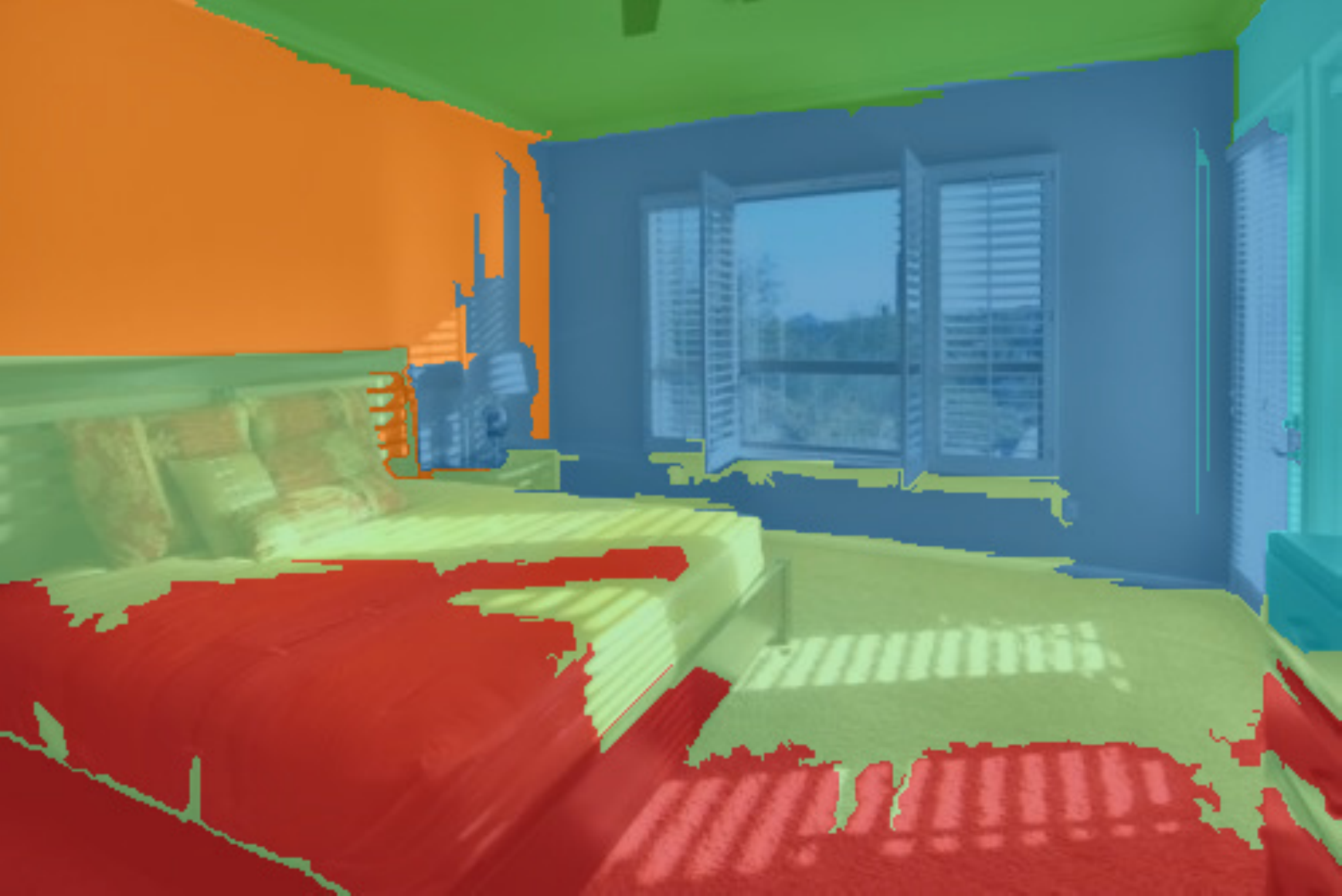}}
	\subfloat[]{\includegraphics[width=.12\textwidth, height=.06\textheight]{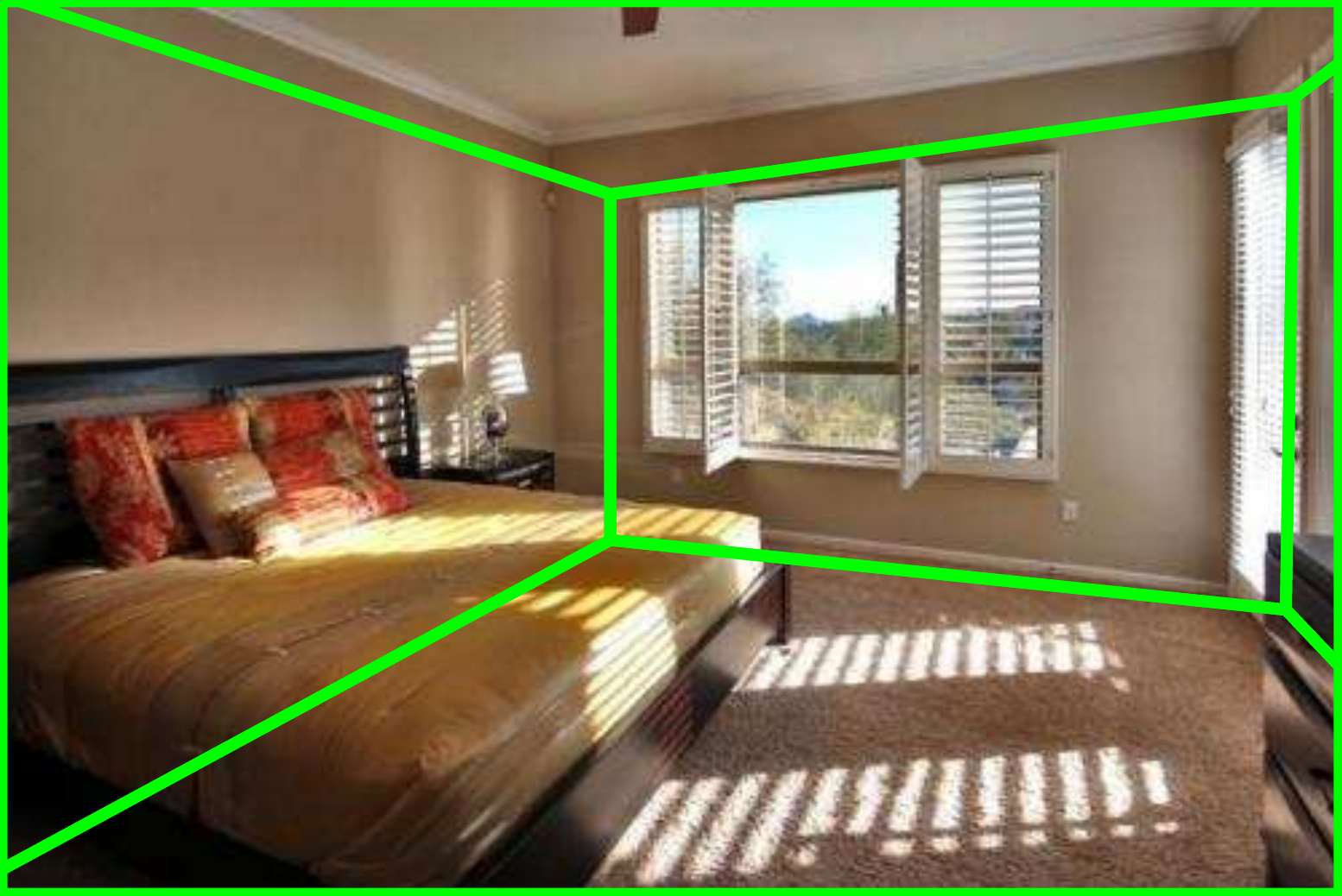}}
	\subfloat[]{\includegraphics[width=.12\textwidth, height=.06\textheight]{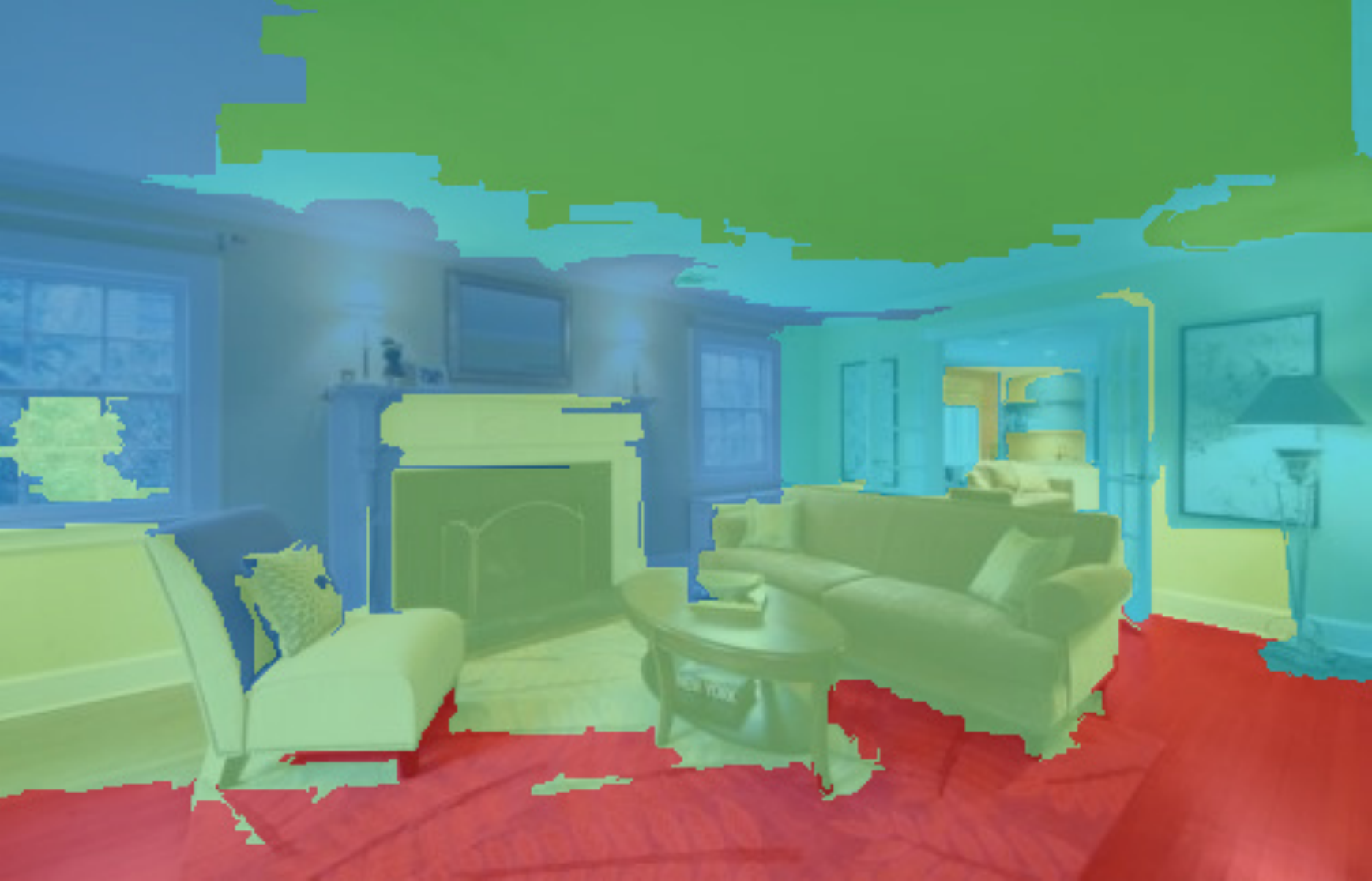}}
	\subfloat[]{\includegraphics[width=.12\textwidth, height=.06\textheight]{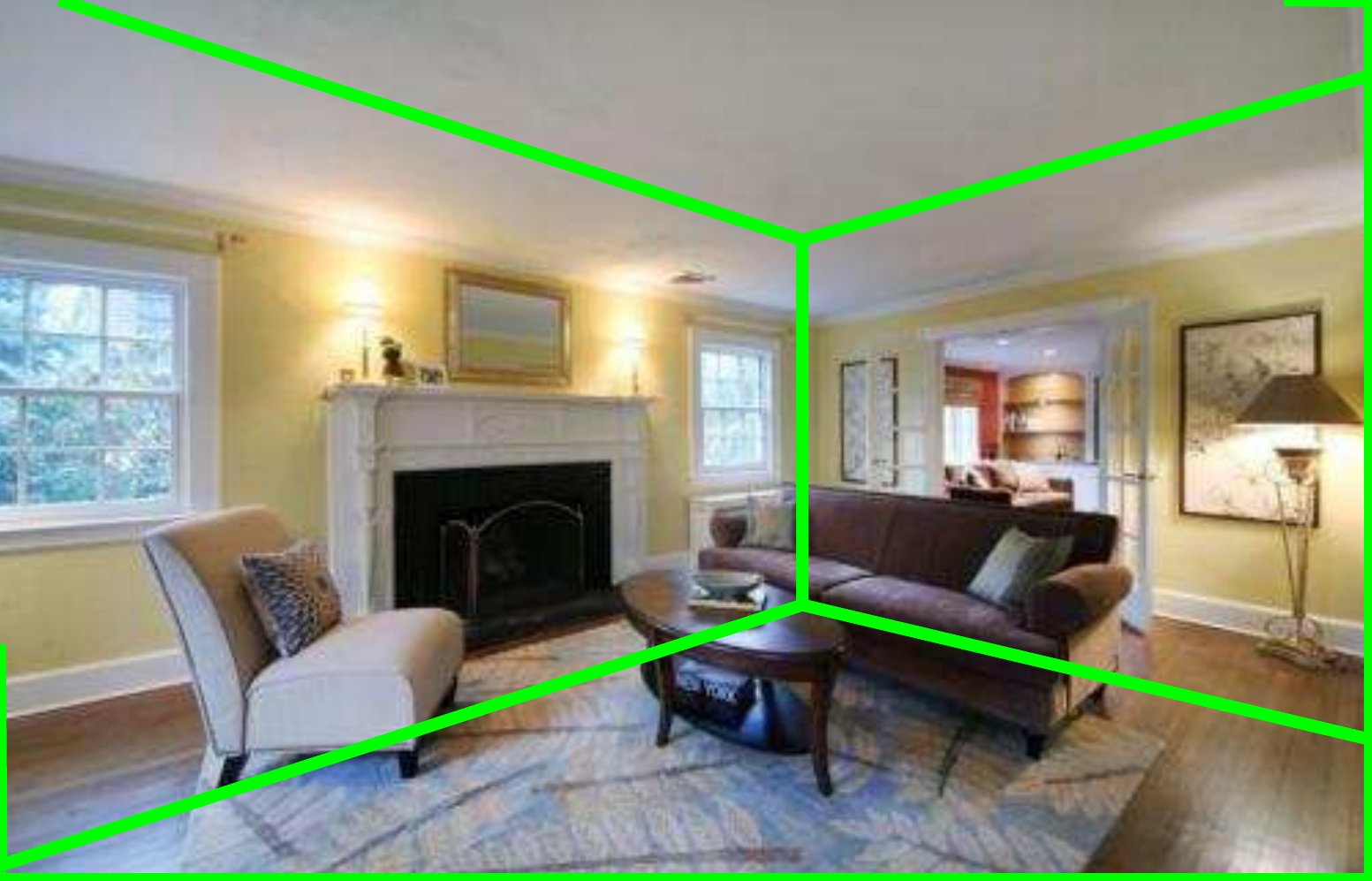}}\quad
	\subfloat[]{\includegraphics[width=.12\textwidth, height=.06\textheight]{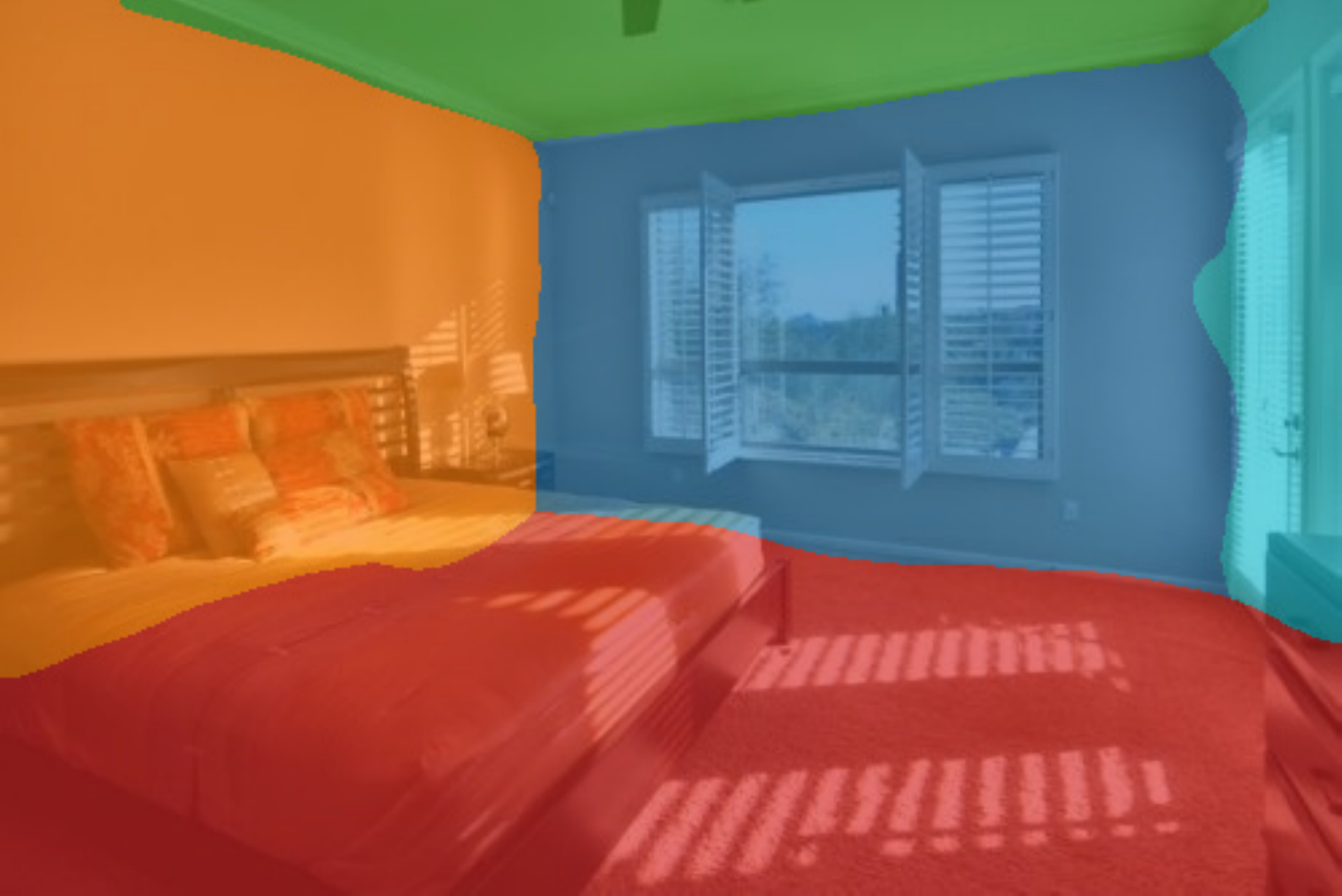}}
	\subfloat[]{\includegraphics[width=.12\textwidth, height=.06\textheight]{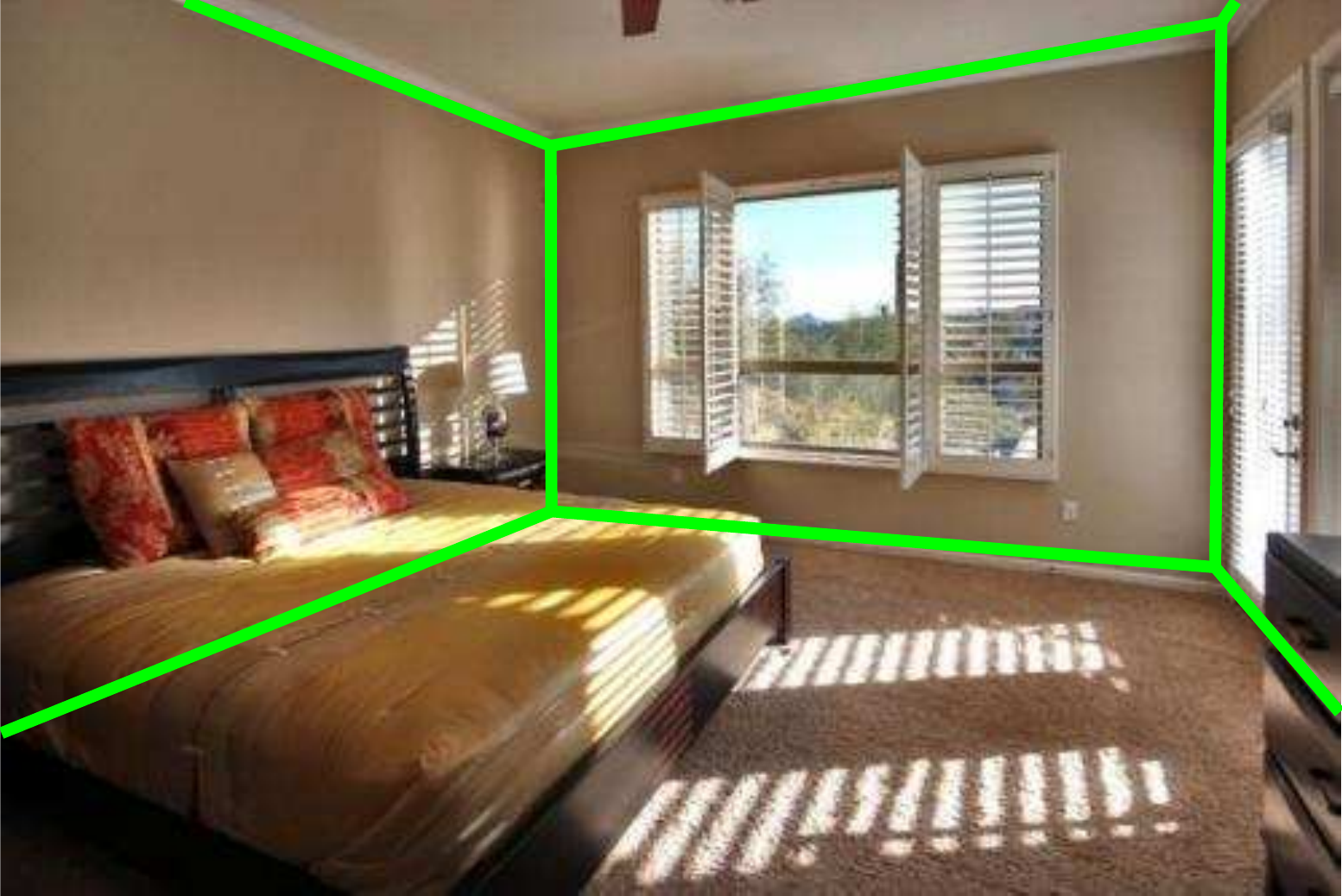}}
	\subfloat[]{\includegraphics[width=.12\textwidth, height=.06\textheight]{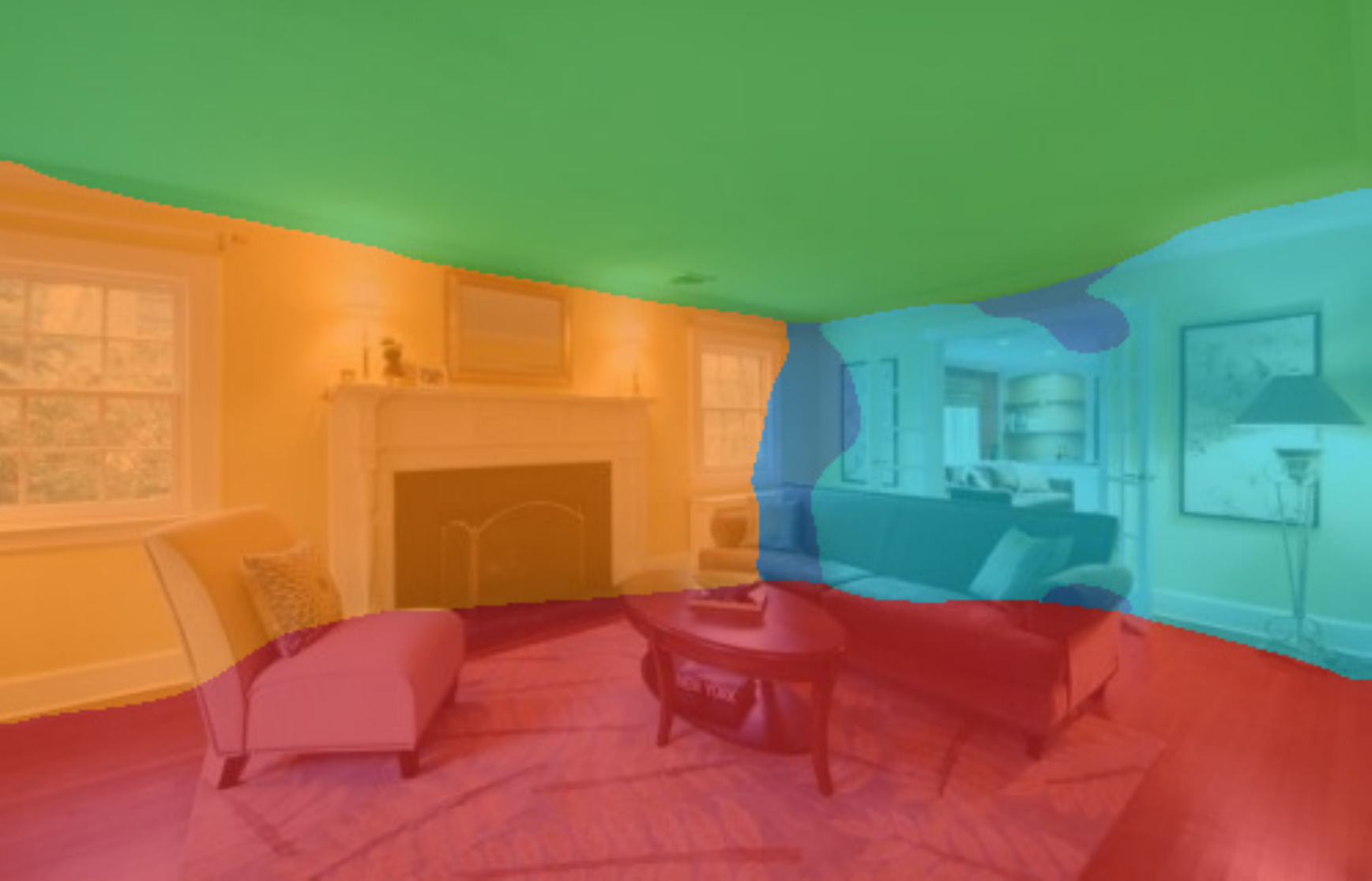}}
	\subfloat[]{\includegraphics[width=.12\textwidth, height=.06\textheight]{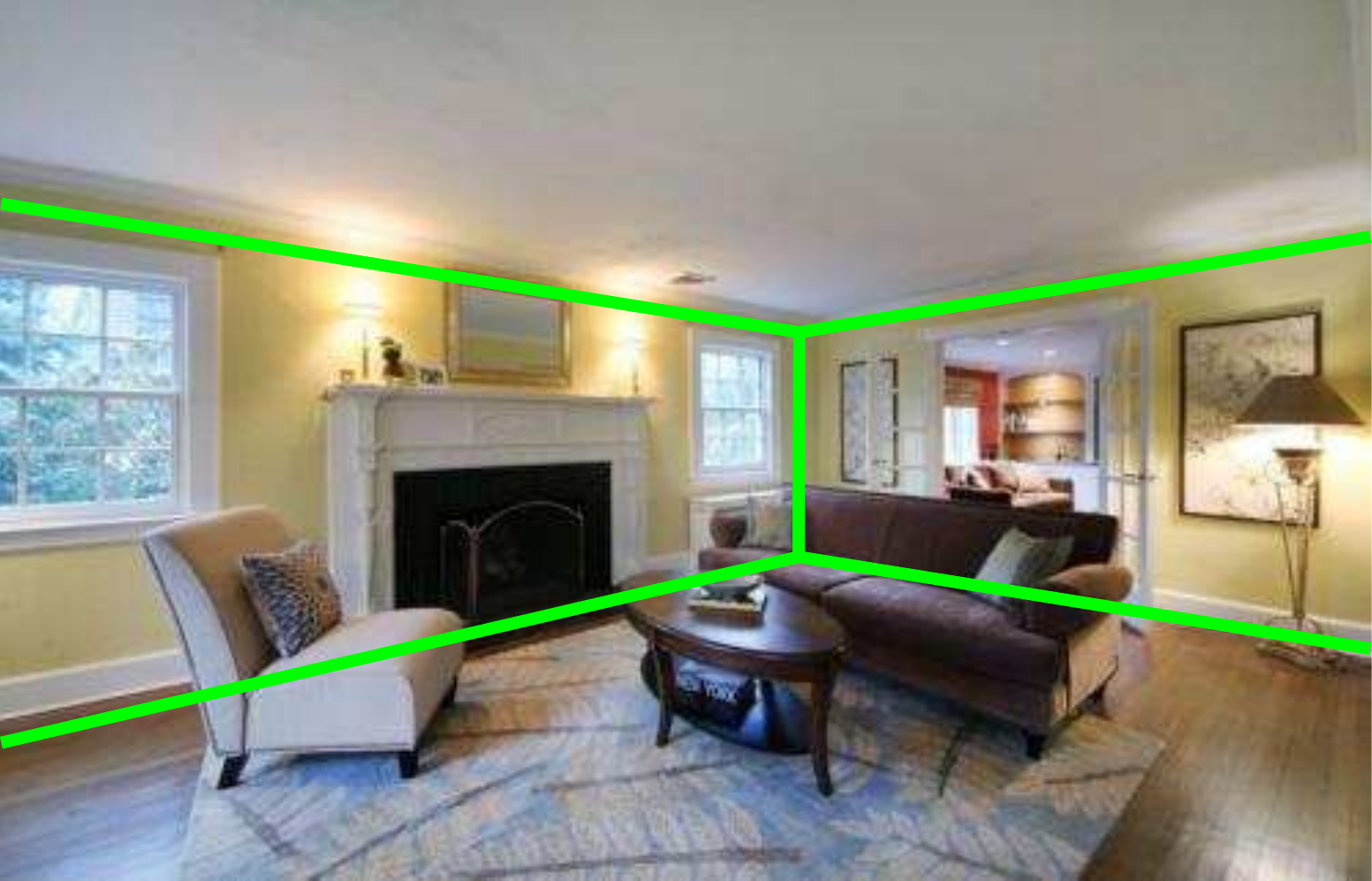}}\quad
	\subfloat[]{\includegraphics[clip, trim=2cm 9cm 2cm 9cm,width=.5\textwidth]{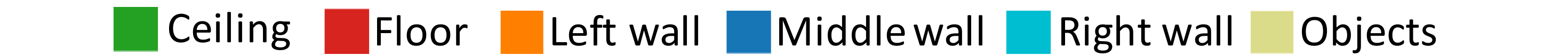}}\quad
	\vspace{-7 mm}
	\caption[3]{\small Geometric feature and room layout estimation. Results from (Row1)~\cite{dclee2009geometric} and (Row2)~\cite{Hedau09}. Bottom row: our results.}
	\label{fig:geometric_feature}
	\vspace{-6 mm}
\end{figure}

\section{Algorithm}
\label{sec:method}
Our approach to reconstructing CAD models from an image (see Figure~\ref{fig:overview}) is based on recognizing objects in the scene, inferring room geometry, and optimizing 3D object poses and sizes in the room to best match synthetic renderings to the input photo.

The proposed approach involves several steps, as follows.  We first fit room geometry, by classifying pixels as being on walls, floor, or ceiling, and fitting a box shape to the result.  In parallel, we detect all of the chairs, tables, sofas, bookshelves, beds, night tables, chest, and windows in the scene using state of the art object detection techniques.  Wherever an object, e.g., a bed, is detected with high confidence, we estimate its 3D pose, by comparing its appearance with renderings of hundreds of beds from many different angles, using a deep convolutional distance metric, trained for this purpose.  Finally, we optimize for the placement of all objects in the reconstructed room by optimizing the difference between the rendered room and the photograph.
Our optimization approach operates on all objects jointly, and thus accounts for inter-object occlusions.

In the remainder of the section, we describe these technical components in detail:  room geometry estimation, object detection, object alignment, and scene optimization.

\subsection{Room Geometry Estimation}
\label{sec:roomGeo}
Humans are adept at interpreting the shape of a room (i.e., positions of walls, ceiling, and floor), even in the presence of significant clutter. Computer vision algorithms have also become increasingly good at this task in the last few years by following a paradigm introduced by Hedau et al.~\cite{Hedau09} and Lee et al.~\cite{dclee2009geometric} in which a set of room shapes are hypothesized (typically 3D boxes), and evaluated using features in the image.

We improve upon previous approaches to room geometry estimation, by adopting an alternative approach for ranking the room 3D box hypothesis using deep convolutional features. Specifically, we train a network that estimates per-pixel surface labels (ceiling, floor, left, middle, and right walls). These features are analogous to the context geometric feature (``support'', ``vertical'', and ``sky'') of~\cite{hoiem2007recovering}.

Unlike~\cite{hoiem2007recovering} that learns the geometry features from hand-designed low level descriptors (e.g., color, texture, and other perspective cues) over superpixels, our method uses an end-to-end deep Fully Convolutional Network (FCN)~\cite{long2015fully}, using VGG~\cite{VGG16} and converting each fully connected layer into a convolutional layer with a kernel covering the entire input region. Finally, the weights are fine-tuned for the pixel-level labeling task. In this work, we produce the output dense score map of size $41 \times 41 \times 5$ given an input image of $321\times321$. We then use upsampling to produce a probability map with the same size of the input image. We trained the FCN network on the annotated indoor scenes in the LSUN dataset~\cite{lsun}.

A key advantage of the FCN-based architecture is that it integrates contextual information over the entire image. Whereas most methods use a ``distractors'' class \cite{Hedau09,mallya15} to remove furniture from consideration, the FCN is able to use furniture as additional context, e.g., using the presence of a bookshelf or bed to infer the likely presence of an adjacent wall. We note that \cite{mallya15} also used a convolutional network, but rather than classifying surface orientations directly as we do, they estimate informative edges in the scene, and employ a second stage to iteratively re-label room surfaces and rank room box estimates.

\subsection{Object Detection}
\label{sec:objectDetPose}

The first step in our furniture modeling pipeline is to detect the presence of objects of interest in the image, and their 2D bounding boxes.  While any number of object detectors can be trained, we focused specifically on the following:  chair, table, sofa, bookshelf, bed, night table, chest, and window.
\begin{figure}[t]
  \captionsetup[subfigure]{labelformat=empty,farskip=-13pt,position=bottom}
  \centering
  \subfloat[]{\includegraphics[width=.235\textwidth]{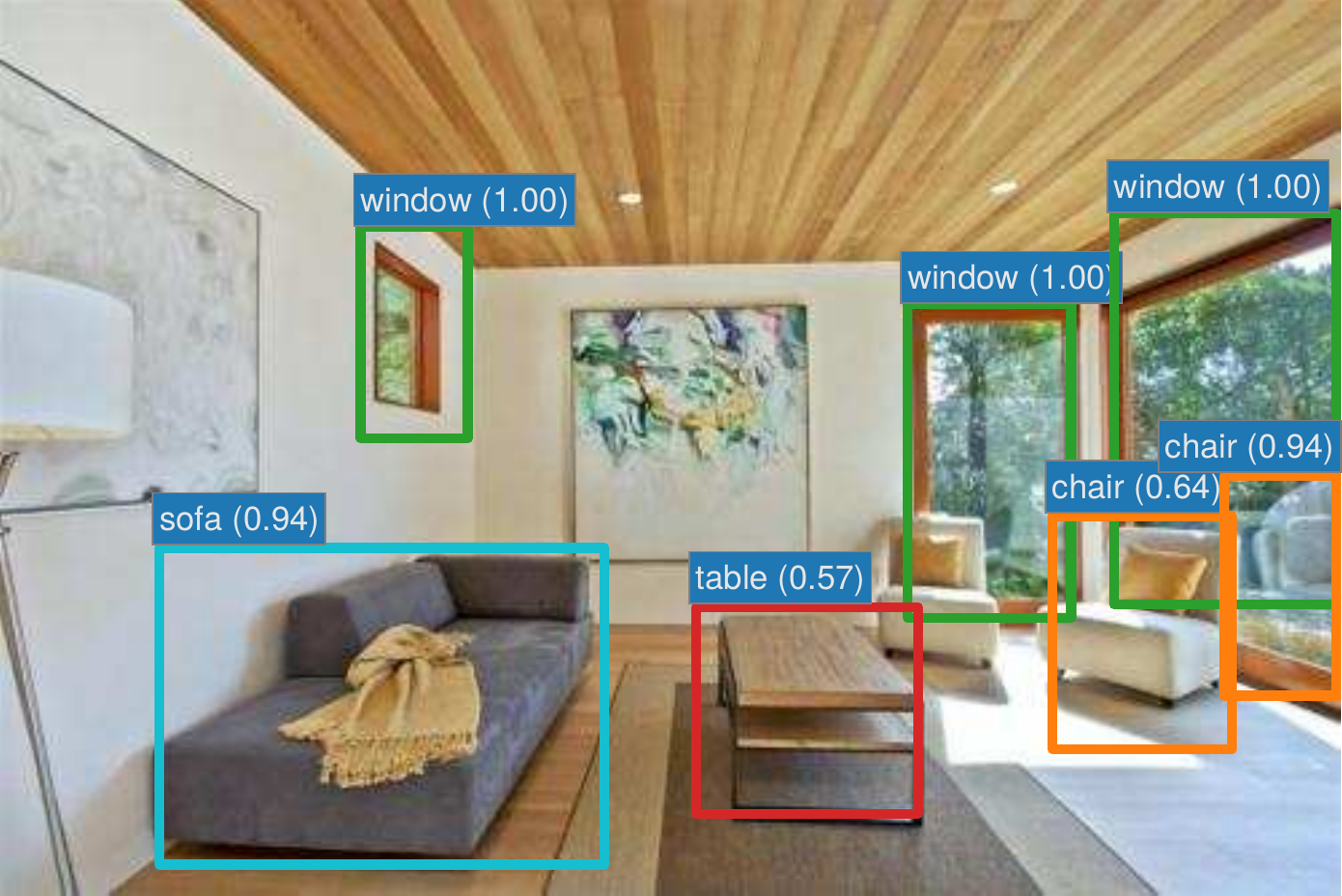}}~\hspace{-0.4em}
  \subfloat[]{\includegraphics[width=.235\textwidth]{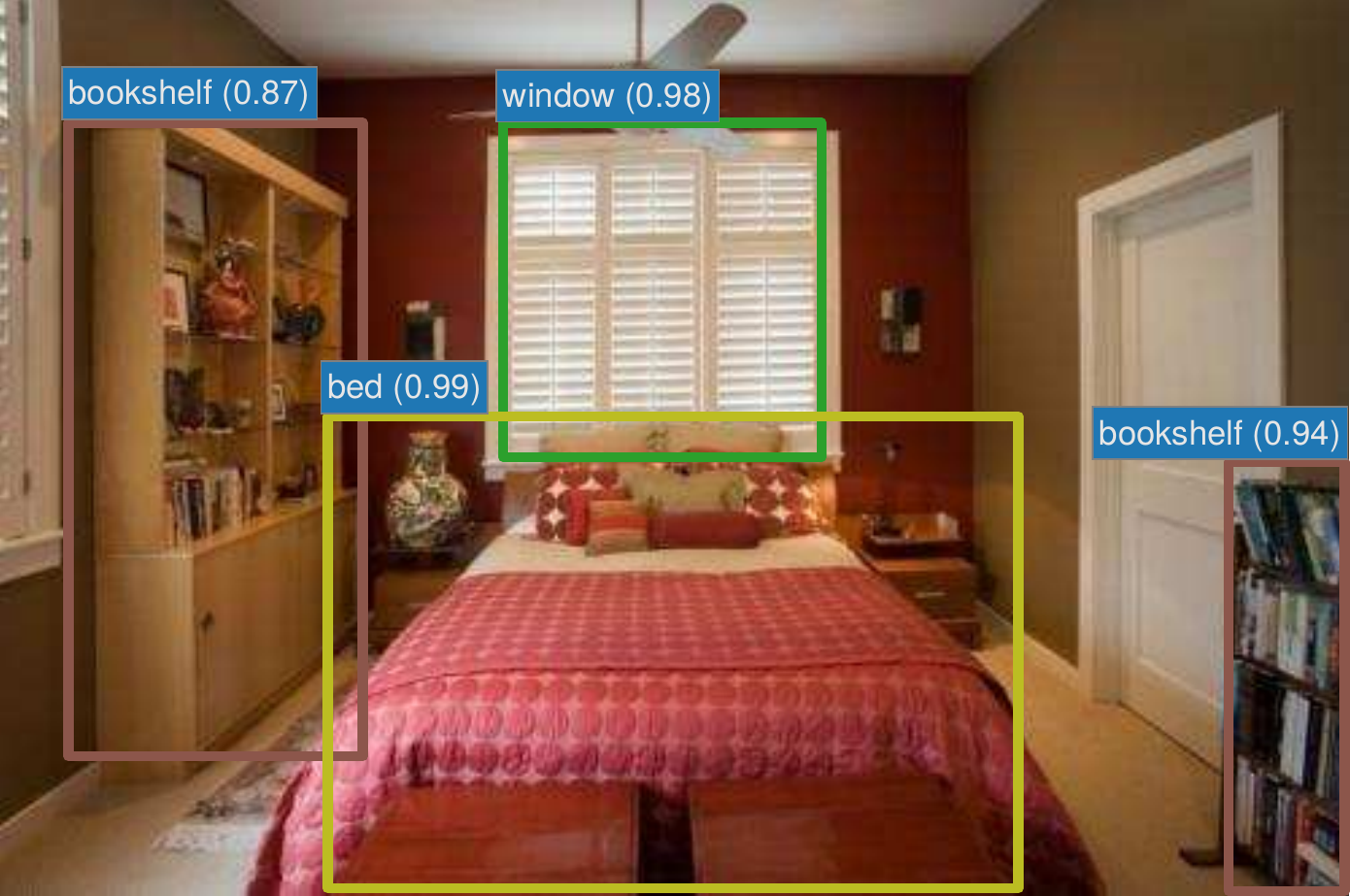}}\quad
  \subfloat[]{\includegraphics[width=.235\textwidth]{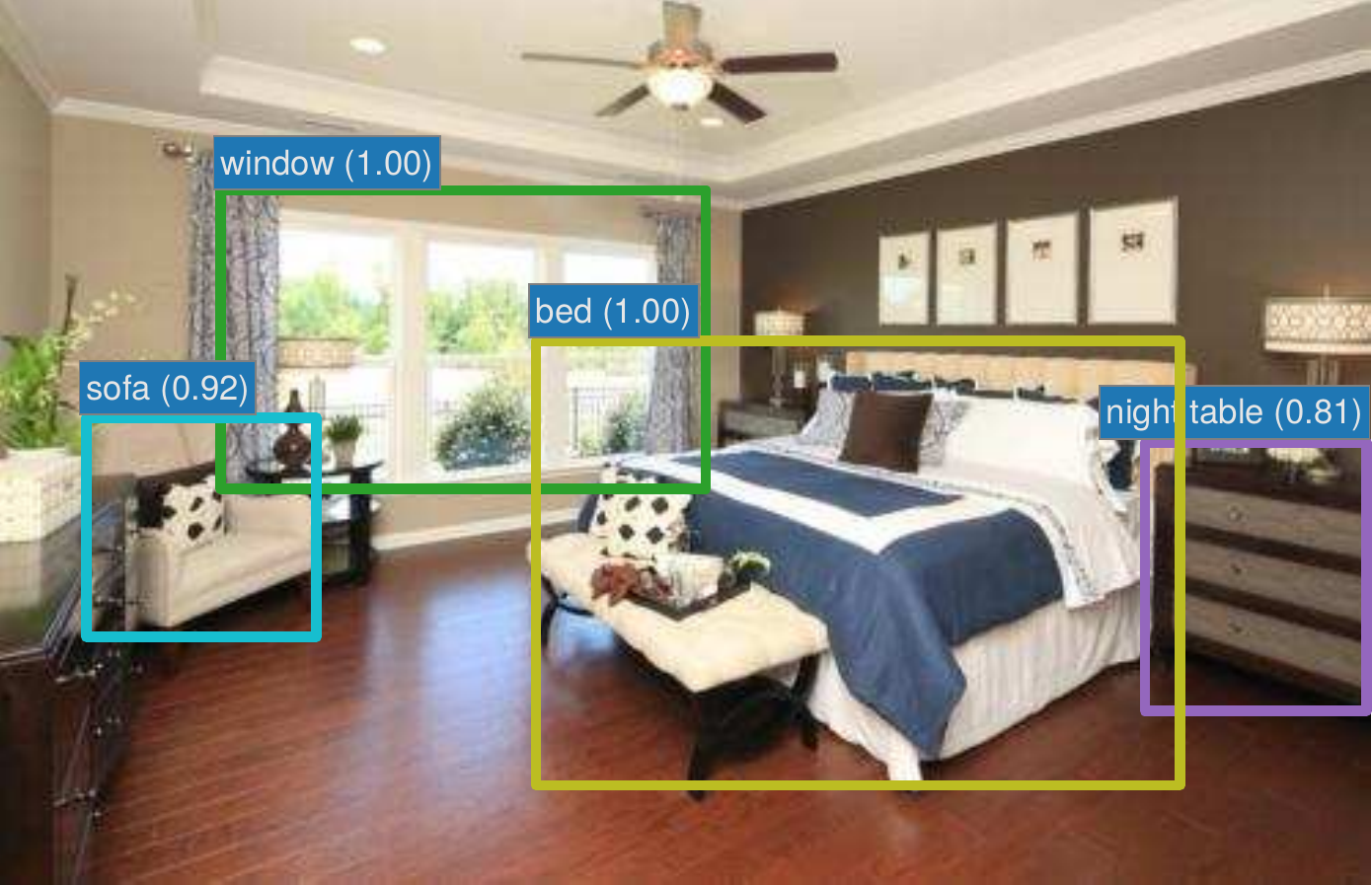}}~\hspace{-0.4em}
  \subfloat[]{\includegraphics[width=.235\textwidth]{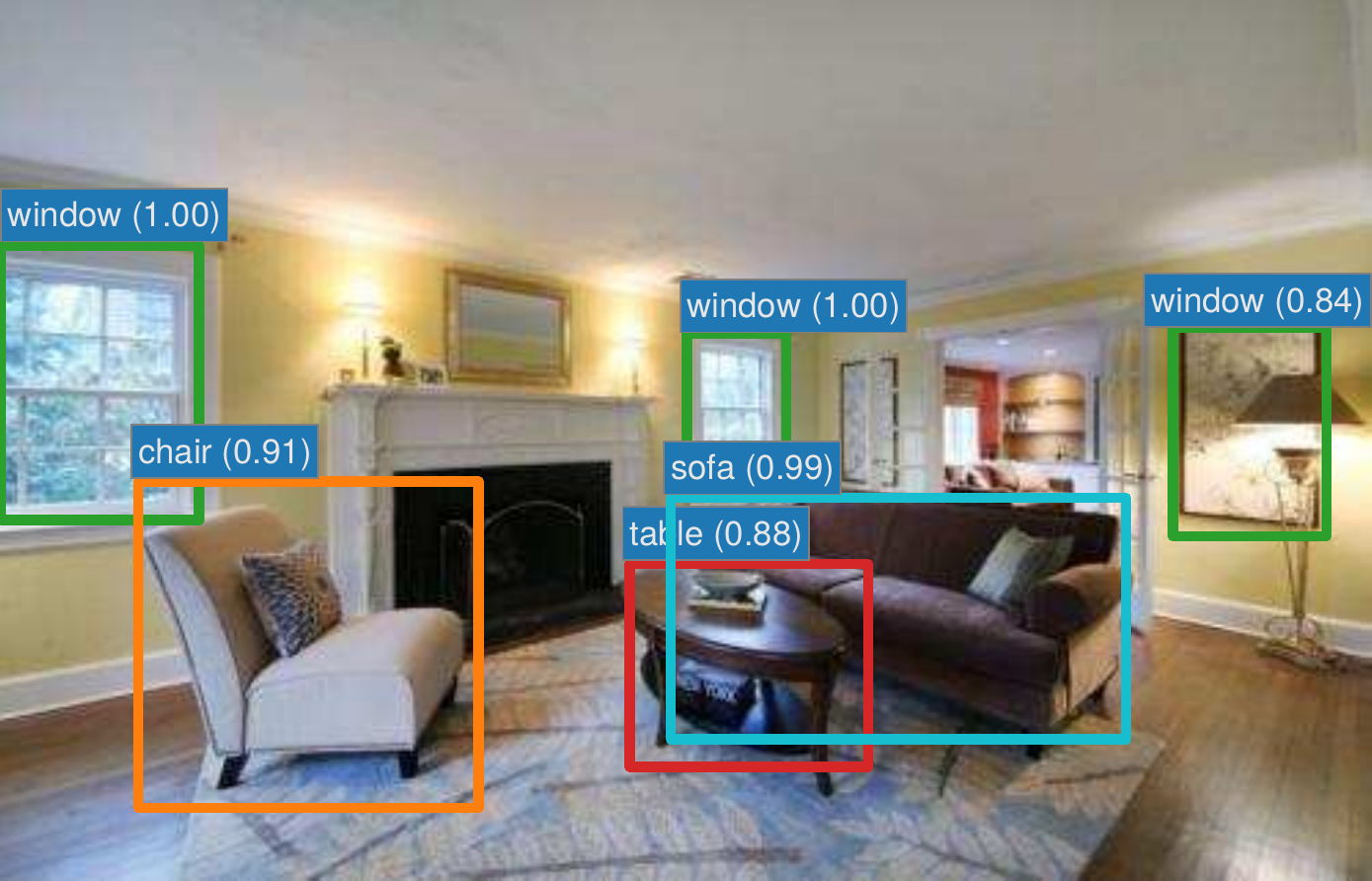}}\quad
  \vspace{-8 mm}
  \caption{\small Object detection result on sample images. Each object category is shown with a different color. The numbers attached to boxes show the probabilities assigned to each detection.}
\label{fig:object_detection}
\vspace{-5 mm}
\end{figure}

Object detection is an area that has seen explosive progress in the last several years, and existing methods work impressively well.  In particular, we use the state-of-the-art Faster-RCNN~\cite{ren2015faster} deep network for detection.  
This network performs two steps to detect objects. First it produces object region proposals, and then it computes the likelihood of each category for the proposed objects using deep convolutional layers. The region proposal layer produces bounding boxes of different scales and aspect ratios. This network is initialized with pre-trained models from large scale object recognition tasks (ILSVRC2012)~\cite{jia2014caffe}. The network weights are then fine-tuned for the object proposal and object detection tasks by minimizing an objective function for a multi-task loss on bounding box regression and object misclassification. The trained network is then able to produce bounding boxes with object categories for any image. The network output also includes an object score which shows the probability of that particular object in the bounding box. Greedy non-maximum suppression (NMS) is used to obtain a single peak detection for each object, remove low scoring detections that overlap with higher scoring object bounding boxes. 

Our Faster-RCNN implementation uses VGG16~\cite{VGG16} architecture. We further fine-tune the weights of this network for the object detection task on our eight furniture categories using three publicly available datasets, namely SUN2012 detection dataset~\cite{sun12}, ImageNet detection challenge dataset~\cite{ILSVRCDetection}, and the window category of Rent3D dataset~\cite{Rent3D}. 
We show detection results on a sample of our images in Figure~\ref{fig:object_detection}.

\begin{figure*}[t]
	\centering
	\includegraphics[width=1.015\textwidth]{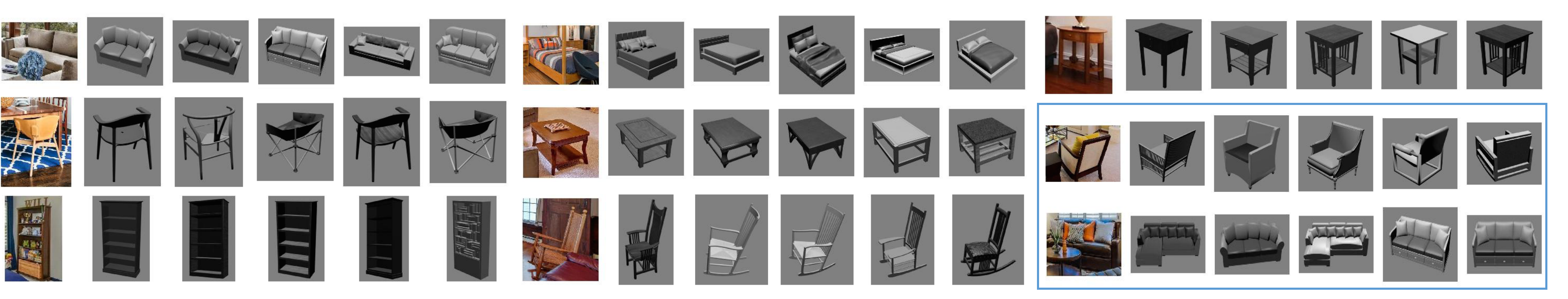}
	\vspace{-8 mm}
	\caption{\small Results of the top five aligned CAD models retrieved for the given object detection bounding box. The retrieved models have similar style and pose with the given object. Last two rows on the right column show failure cases: (Row1) visual feature confusion between different poses of the chair, and (Row2) heavy occlusion of sofa by table has made it visually similar to an L-shaped sofa.}
	\label{fig:object_pose_estimation}
	\vspace{-4 mm}
\end{figure*}

\subsection{CAD Model Alignment}
\label{sec:CAD Model Alignment}
The object detection results from Section~\ref{sec:objectDetPose} identify the presence of a ``chair'' (e.g.,) in a certain region of the image with high probability.  Now we wish to determine what {\em kind} of chair it is, its shape, and approximate 3D pose.

Inspired by~\cite{aubry2014seeing}, we solve this retrieval problem by searching for 3D models that are most similar in appearance to the detected objects in the image. Specifically, we consider all 3D models in the ShapeNet repository~\cite{shapenet} associated with our object categories of  interest, i.e., chair, table, sofa, bookshelf, bed, night table, chest, yielding 9193 models in total. Each 3D model is rendered to 32 viewpoints, consisting of 16 uniformly sampled azimuth angles and two elevation angles (15 and 30 degrees above horizontal).

Robust comparison of photos with CAD model renderings is not straightforward; simple norms like L2 do not work well in practice, due to differences in shape, appearance, shading, and the presence of occluders. We achieve good results, once again, by using convolutional nets (see Figure~\ref{fig:object_pose_estimation}); we compute deep features for each of the rendered images and the detected image bounding boxes and use cosine similarity as our distance metric. More specifically, we use the convolution filter response in the ROI-pooling layer of the fine-tuned Faster-RCNN network~\cite{ren2015faster} described in Section~\ref{sec:objectDetPose}. A benefit of using the ROI-pooling layer is that the length of its feature vector does not depend on the size and the aspect ratio of the bounding box, thus avoiding the need for non-uniform rescaling (a source of artifacts in general). Choosing the rendering that best matches each image object detection yields an estimate both for the best-matching CAD model and its approximate 3DOF orientation.

\subsection{Object Placement in the Scene}
\label{sec:placement}
Equipped with a set of CAD models and their approximate orientations, we now wish to place them in the reconstructed room. This placement need not be exact, as we will further optimize it in a subsequent step, but should be a reasonable initial estimate. To this end, we first estimate the intrinsic camera parameters $(K)$ and camera rotation $(R)$ with respect to the room space using three orthogonal vanishing points~\cite{Hedau09}, and choose one of the visible room corners as the origin of the world coordinate system. If none of the corners are visible, we choose the origin to be the intersection of the visible wall edge with the floor.

The ShapeNet 3D models are normalized with a bounding box corresponding to a unit cube. Based on the alignment procedure from Section \ref{sec:CAD Model Alignment}, we can determine the input photo pixel locations corresponding to each of the eight corners of this cube. We can find the object location and scale in the $x$ and $y$ (parallel to ground plane) directions by intersecting the ground plane with the ray casted from the camera center through the input image pixels corresponding to the bottom four corners of the aligned CAD model cube. To compute the object scale along the $z$ axis, we compute the ratio between the length of the four vertical edges of the projected cube and the length of those edges from the ground plane to the intersection of those lines with the horizontal vanishing line. Note that the height of the vanishing line is equal to the camera height. 

We treat windows as a special case, as they are attached to walls instead of floor. To place windows, we find the intersection of the window bounding box from object detection with each of the walls and assign the window to the wall with which it has the largest overlap. The window's detected bounding box in the image back-projects to a quadrilateral on the assigned wall. The pose and location of window is determined by the largest axis-aligned rectangle on the wall plane contained within that quadrilateral.

\begin{figure*}[t!]
  \captionsetup[subfigure]{labelformat=empty,farskip=-12pt,position=bottom}
  \centering
  \subfloat[]{\includegraphics[width=.14\textwidth,height=.10\textwidth]{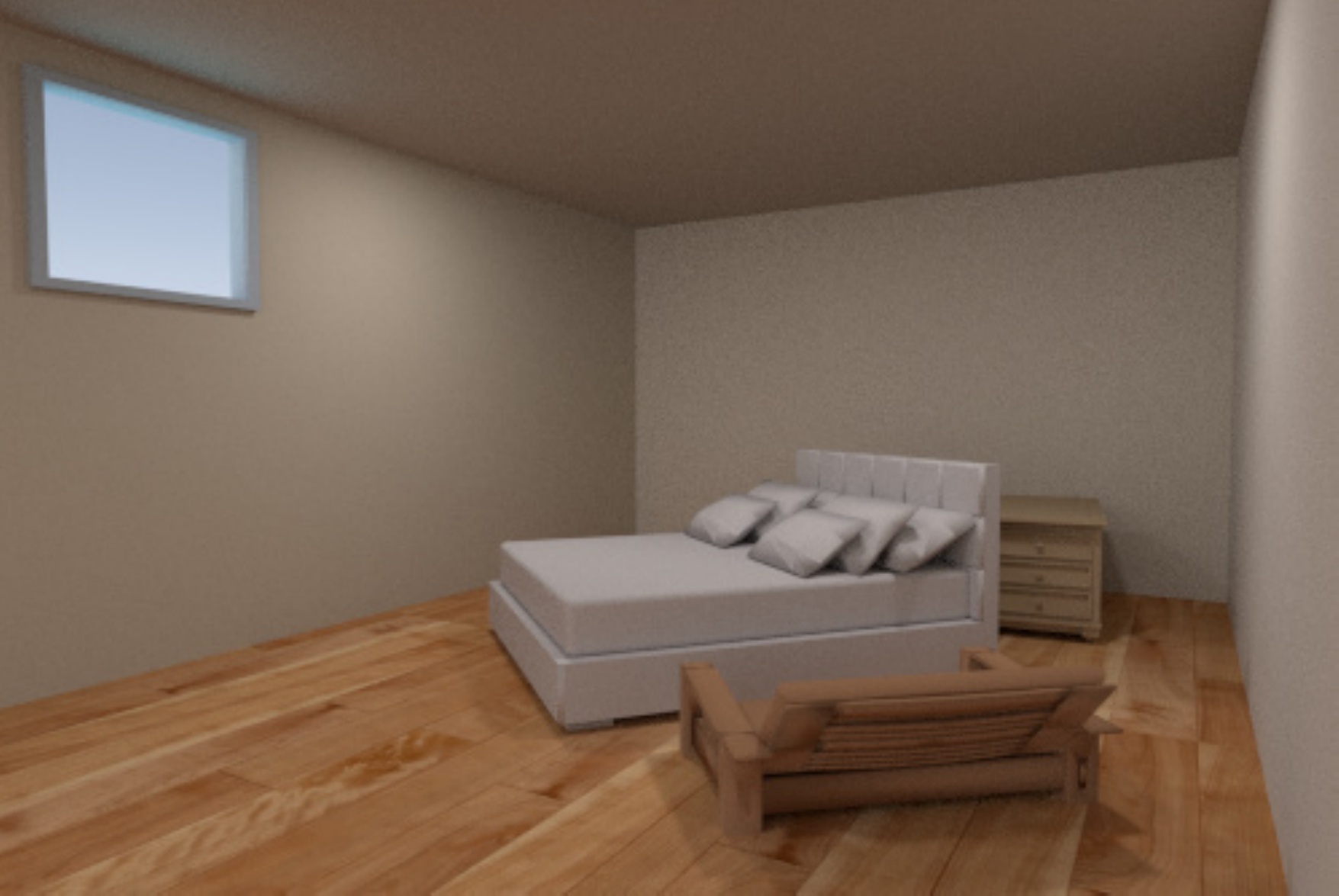}}~\hspace{-0.3em}
  \subfloat[]{\includegraphics[width=.14\textwidth,height=.10\textwidth]{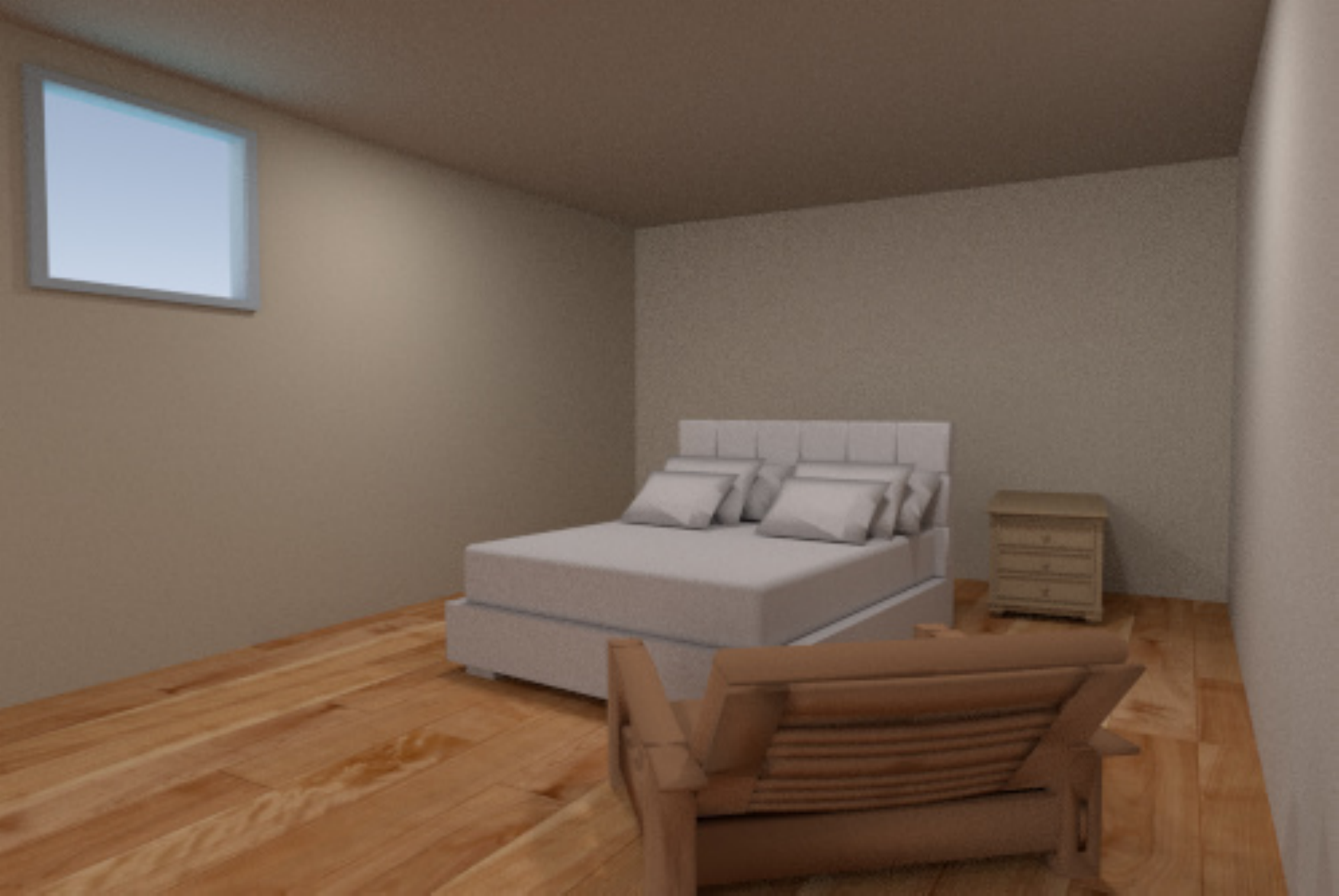}}~\hspace{-0.3em}
  \subfloat[]{\includegraphics[width=.14\textwidth,height=.10\textwidth]{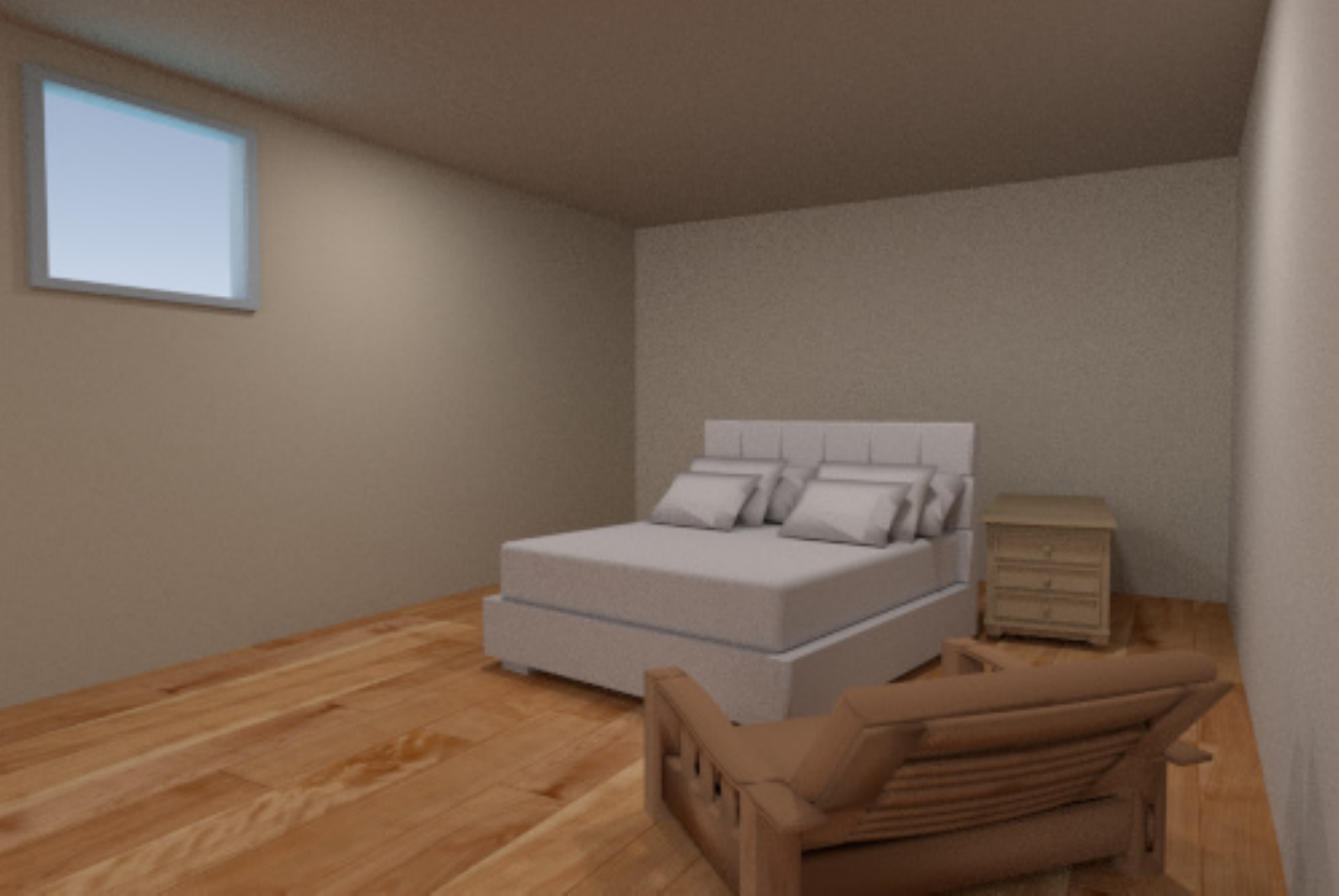}}~\hspace{-0.3em}
  \subfloat[]{\includegraphics[width=.14\textwidth,height=.10\textwidth]{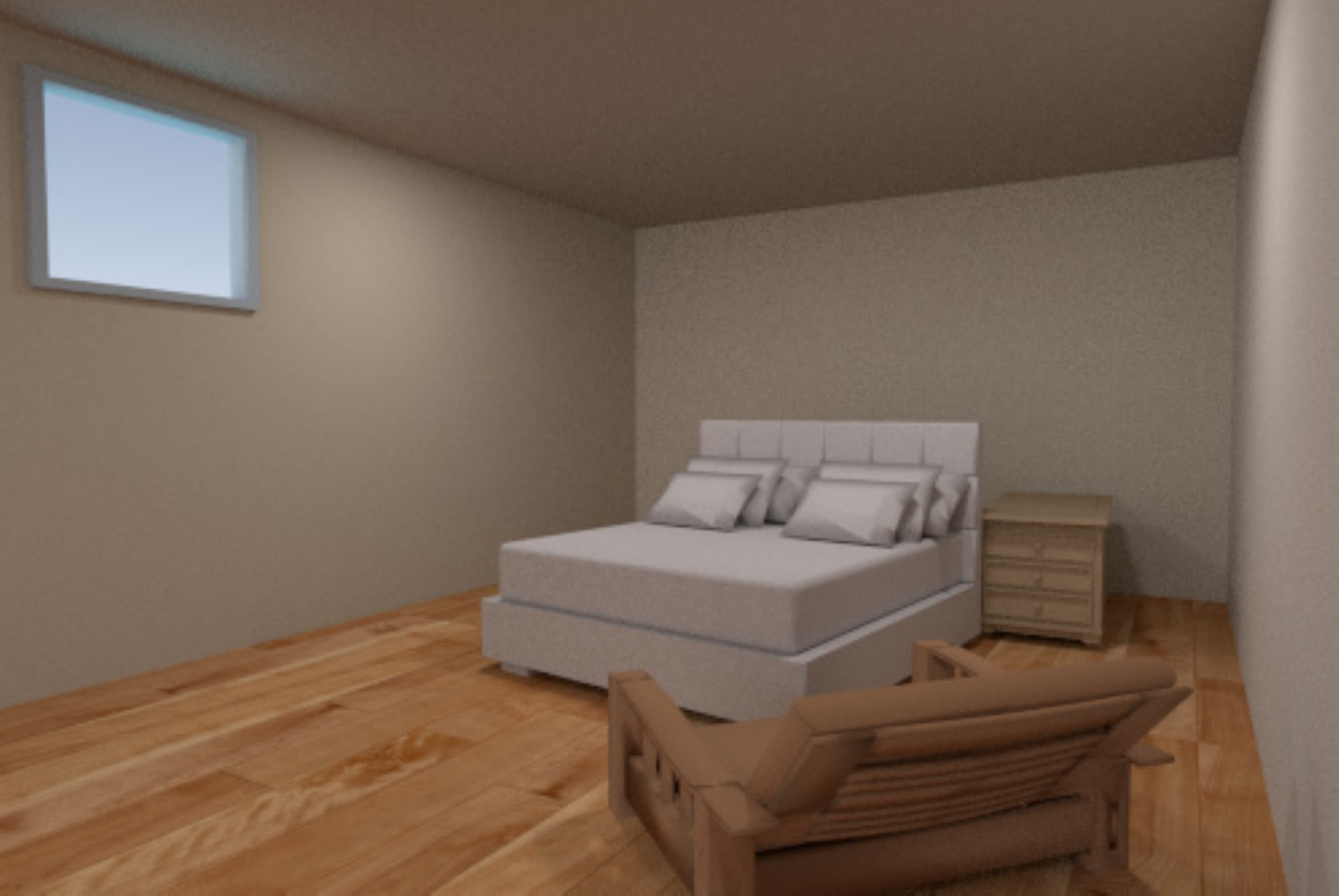}}~\hspace{-0.3em}
  \subfloat[]{\includegraphics[width=.14\textwidth,height=.10\textwidth]{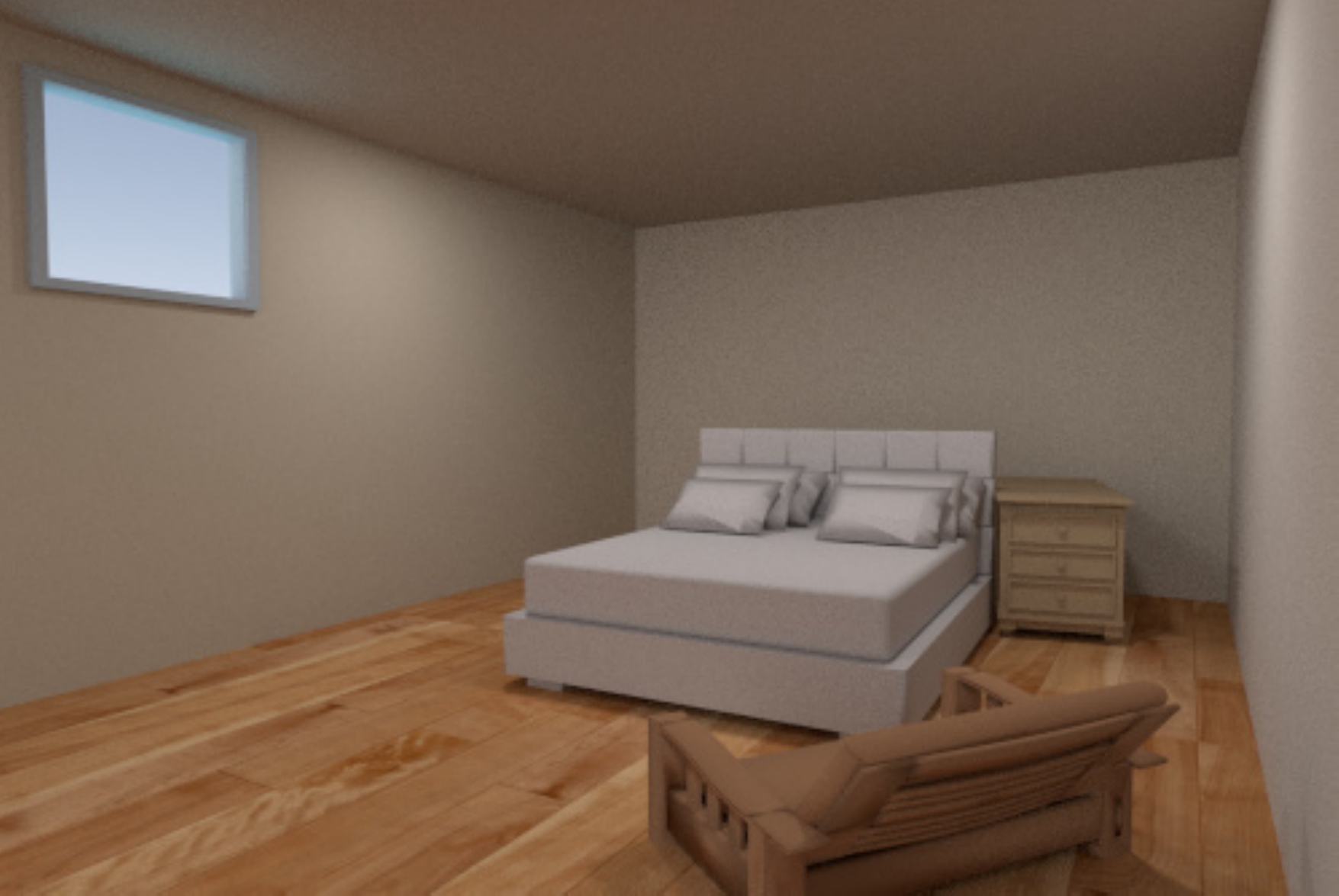}}~\hspace{-0.3em}
  \subfloat[]{\includegraphics[width=.14\textwidth,height=.10\textwidth]{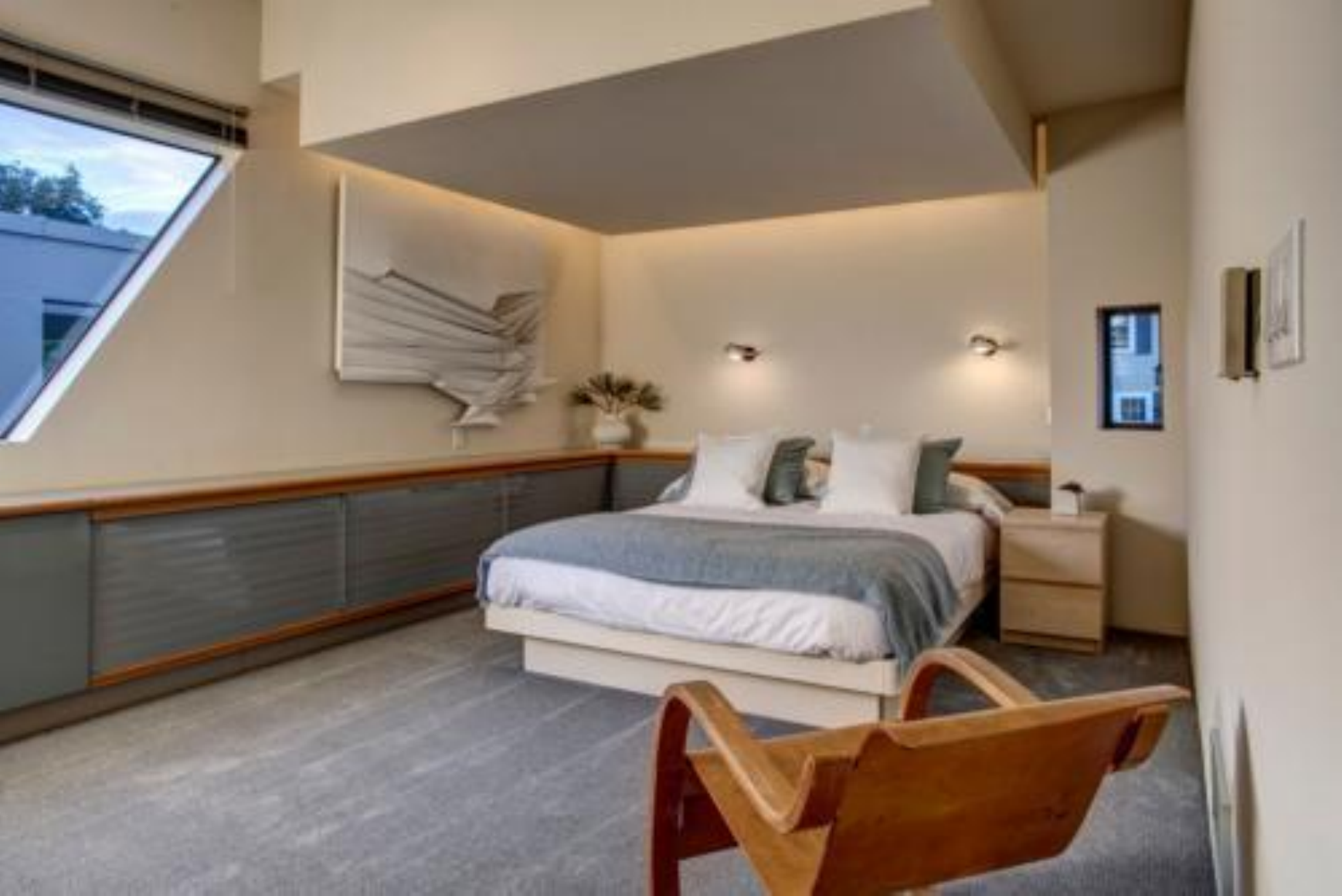}}~\hspace{-0.3em}
  \subfloat[]{\includegraphics[width=.14\textwidth,height=.10\textwidth]{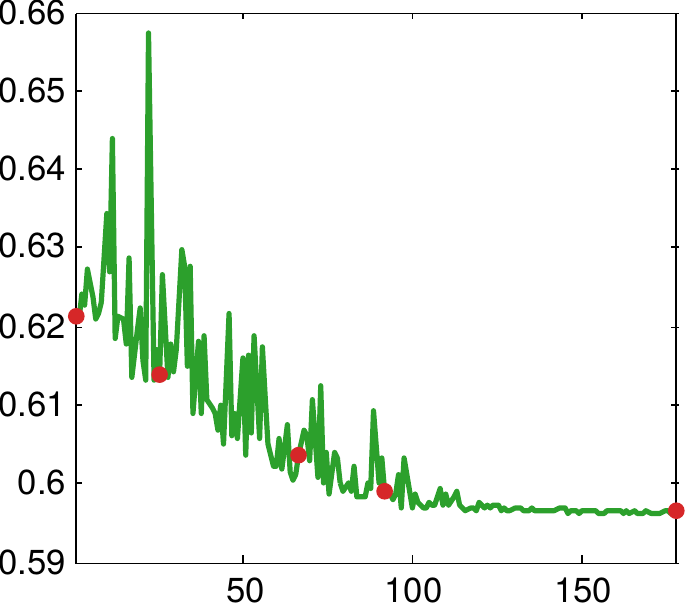}}\quad

  \subfloat[]{\includegraphics[width=.14\textwidth,height=.10\textwidth]{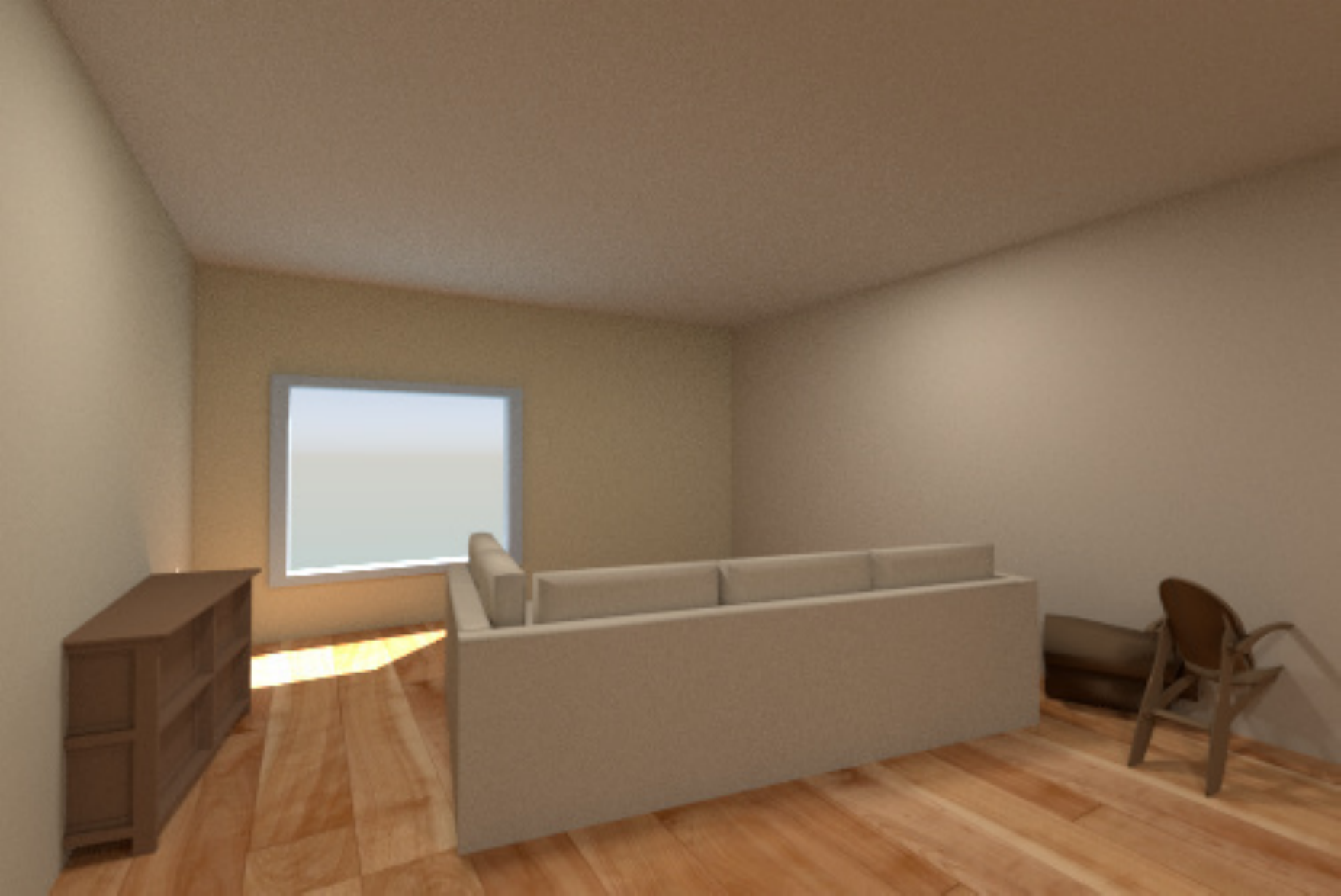}}~\hspace{-0.3em}
  \subfloat[]{\includegraphics[width=.14\textwidth,height=.10\textwidth]{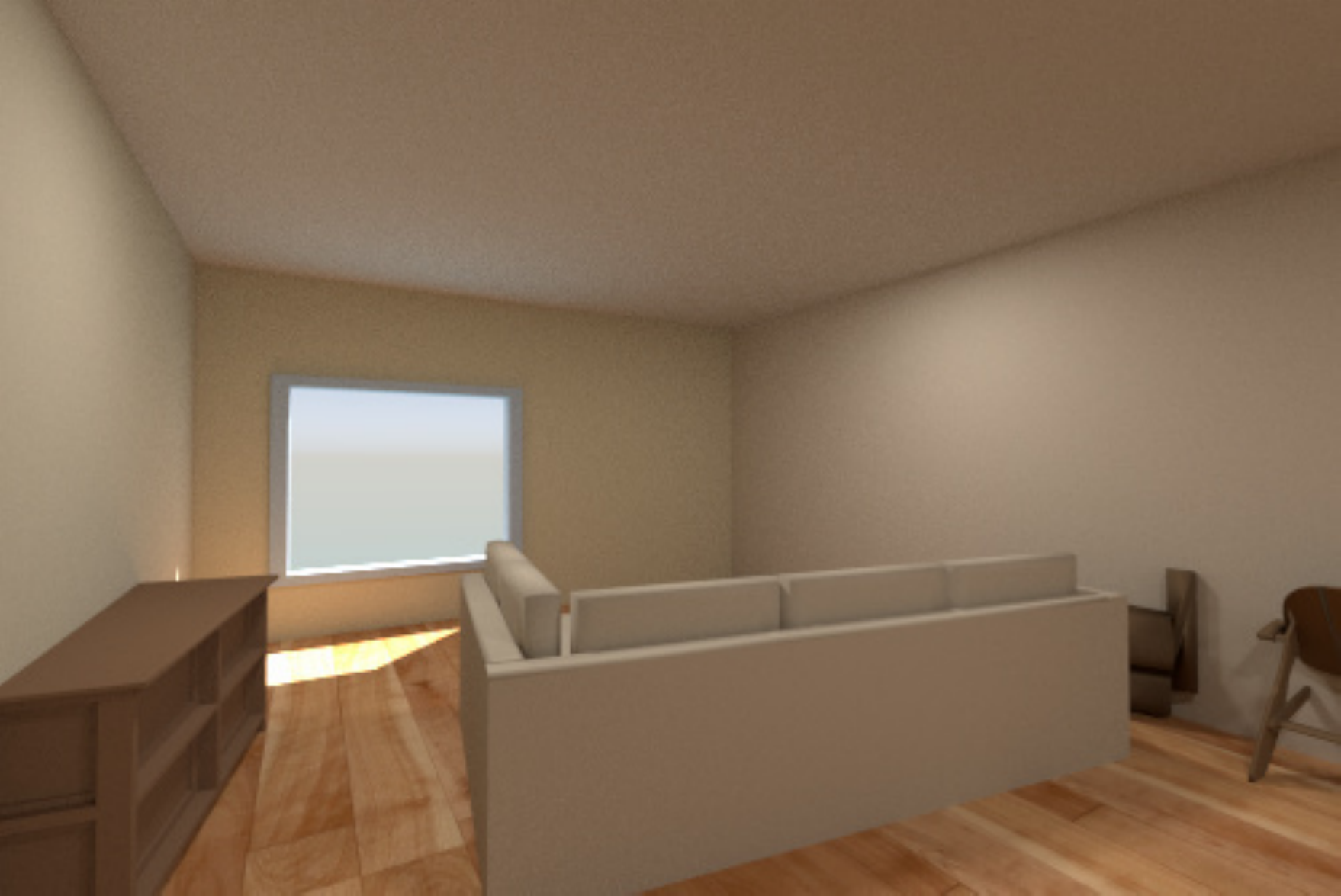}}~\hspace{-0.3em}
  \subfloat[]{\includegraphics[width=.14\textwidth,height=.10\textwidth]{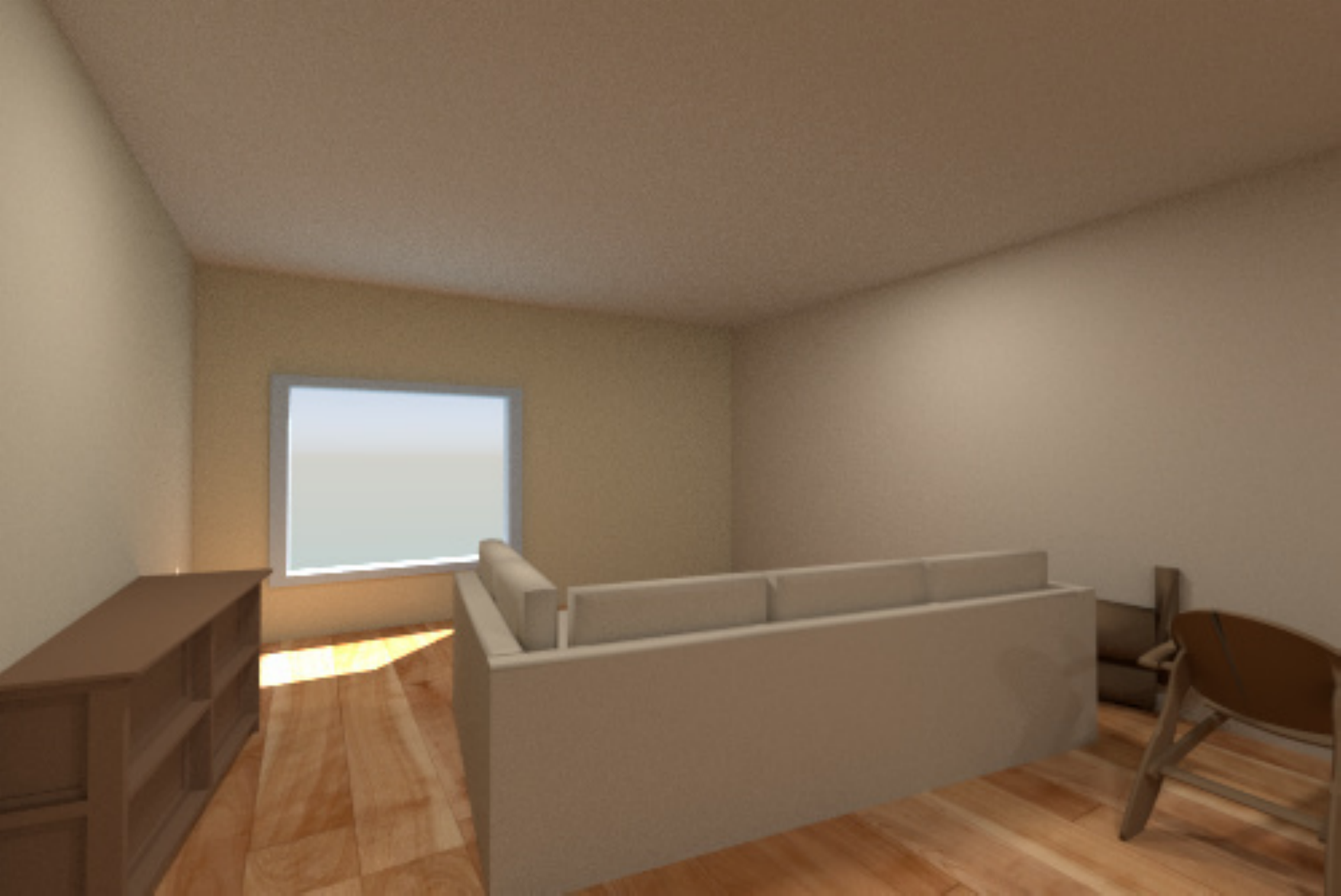}}~\hspace{-0.3em}
  \subfloat[]{\includegraphics[width=.14\textwidth,height=.10\textwidth]{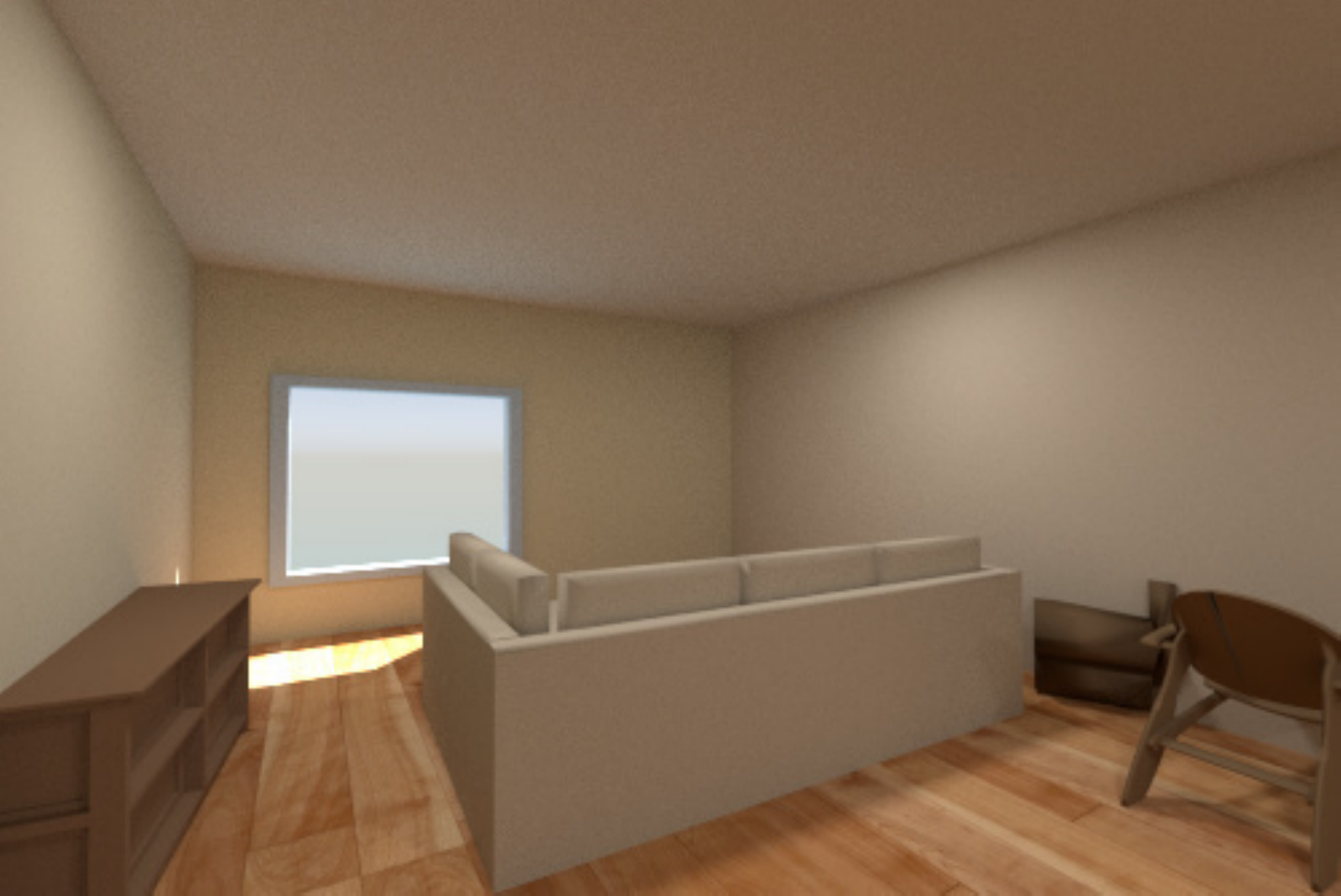}}~\hspace{-0.3em}
  \subfloat[]{\includegraphics[width=.14\textwidth,height=.10\textwidth]{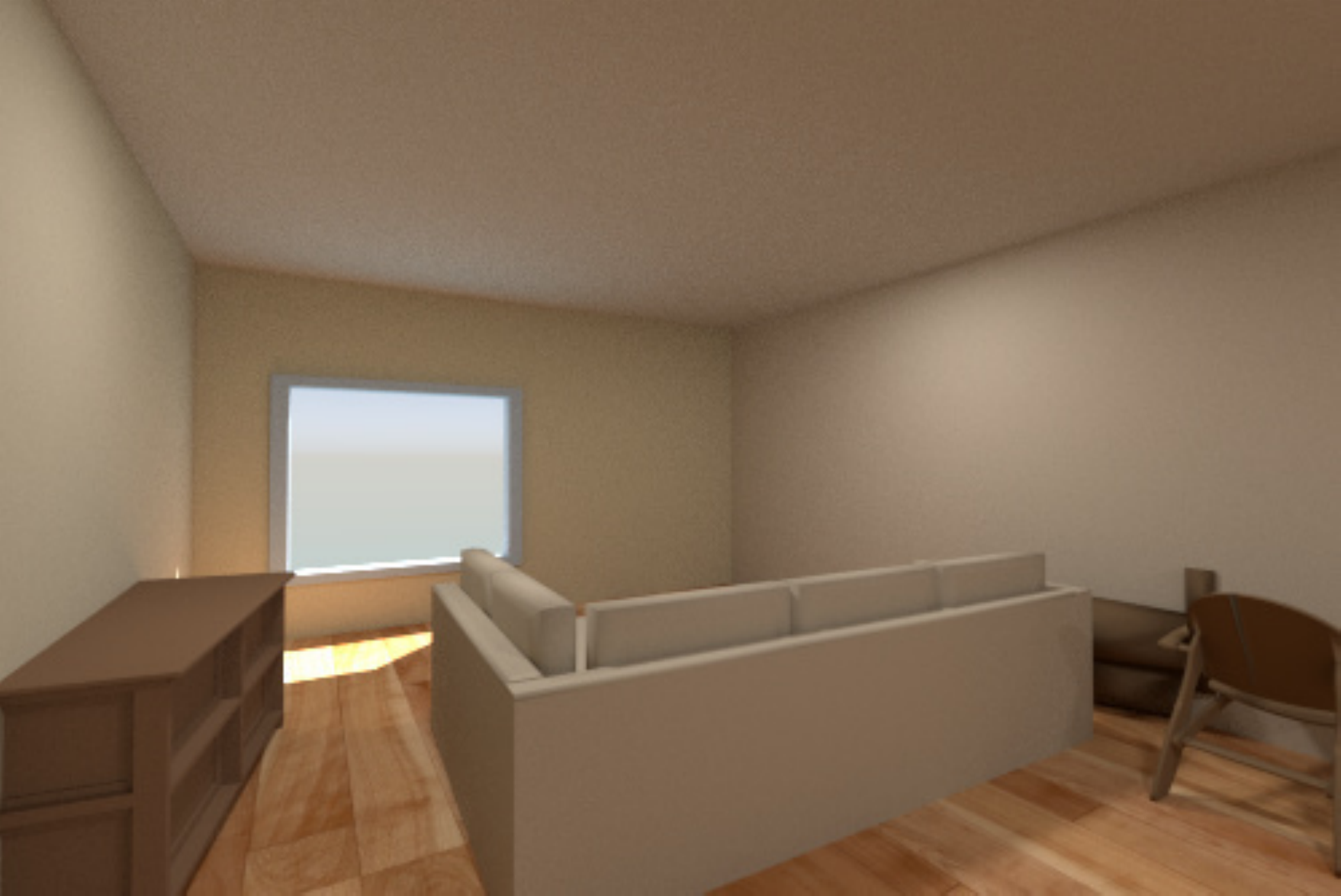}}~\hspace{-0.3em}
  \subfloat[]{\includegraphics[width=.14\textwidth,height=.10\textwidth]{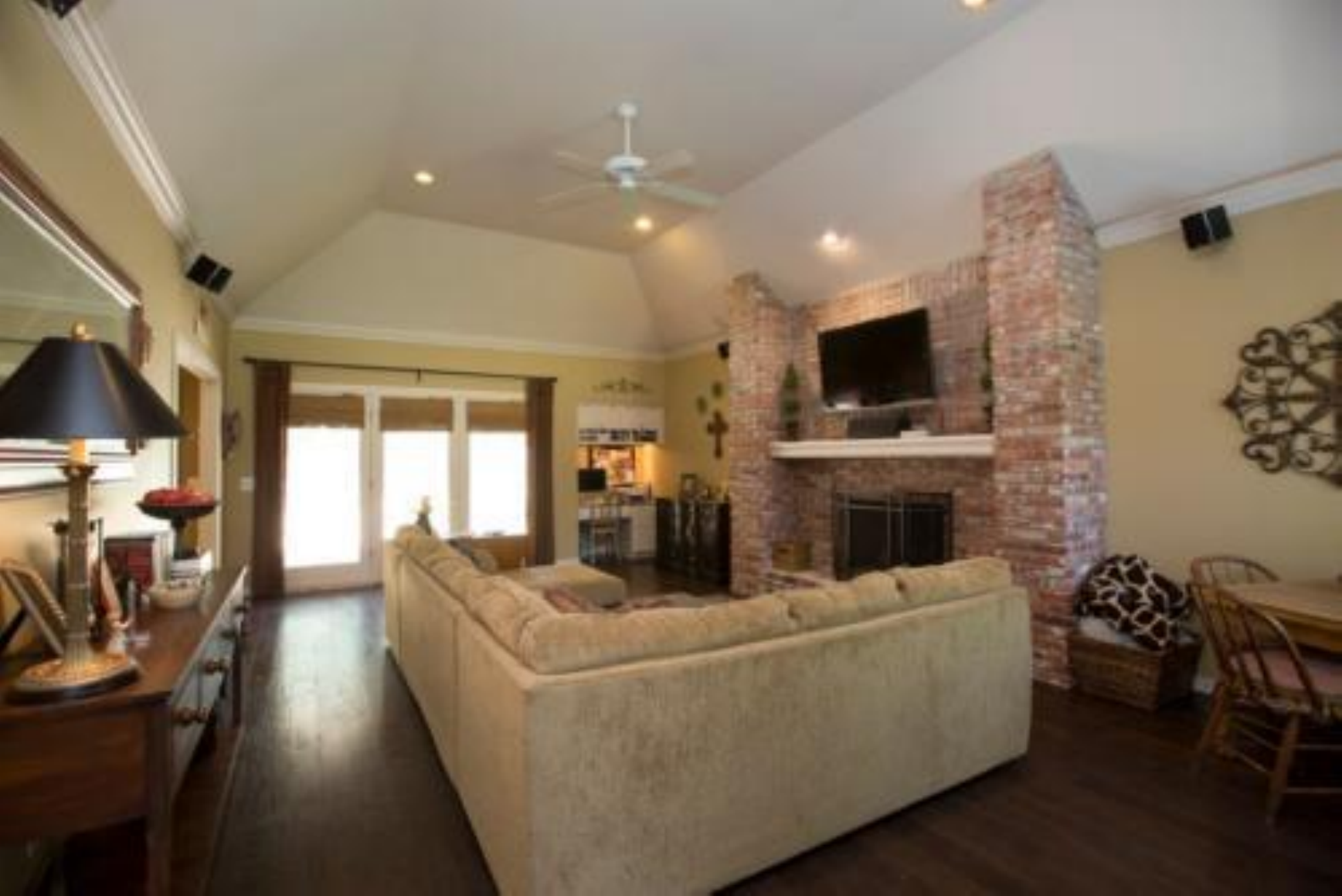}}~\hspace{-0.3em}
  \subfloat[]{\includegraphics[width=.14\textwidth,height=.10\textwidth]{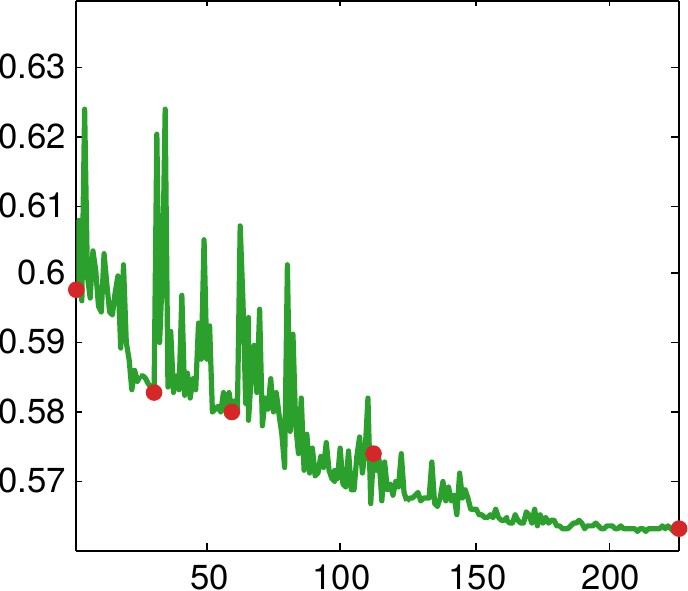}}\quad
  \subfloat[]{\includegraphics[width=1.007\textwidth]{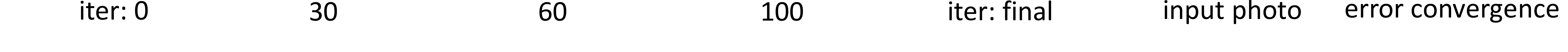}}\quad
  \vspace{-11 mm}
  \caption{\small Results of the joint scene optimization step. (Column 1) The initial object placement in the scene. (Columns 2-5) Rendering of the scene in sample iterations during optimization. (Column 5) The last iteration of optimization. (Last column) The objective function error and the optimization convergence. The objective function minimizes dis-similarity between the real and the rendered image. Red dots show the sample iterations that are shown above.}
\label{fig:optimization_convergence}
  \vspace{-5 mm}
\end{figure*}

\subsection{Scene Optimization via Render and Match }
\label{sec:sceneOptimization}

The placement procedure in Section~\ref{sec:placement}
is sensitive to several sources of error including the quantization of object orientations, ground plane misregistration, occlusions, and other factors, which can lead to erroneous estimates of object pose and scale. We therefore introduce an optimization in which the configurations of all objects in the scene are jointly aligned. A benefit of this procedure is that it properly accounts for inter-object occlusions, and yields more accurate estimates for object location, scale, and orientation.

After estimating the 3D room geometry and the initial placement of the objects in the scene, we refine our object placements by optimizing the visual similarity of the rendered scene with that of the input image. To this end, we solve an optimization problem where the variables are the 3D object configurations in the scene and the objective function is the minimization of the cosine distance between the convolutional features obtained from the camera view rendered scene and the input image.

More formally, suppose we detect objects $\{O_1,...,O_k\}$ in the scene. The placement of each object $O_i$ is represented by its $(x,y,z)$ location, scale along the $x$, $y$ and $z$ axis as well as the rotation. The variables for all $N$ objects are concatenated into a $7N$ parameter vector. Given a parameter vector, we can generate the rendered image of the scene, denoted $I^*$. The cost function used in our optimization tries to minimize the cosine distance between $I^*$ and the original input image $I$:
\vspace{-4 mm}

\begin{equation}
 \min \Phi(I^*,I) = \frac{1}{|\mathcal{C}|} \sum_{C_i \in \mathcal{C}}{1-\frac{C_i(I^*)\cdot C_i(I)}{\parallel C_i(I^*)\parallel \parallel C_i(I)\parallel}}
\end{equation}

We model the feature vector of an image by using the outputs of all convolutional layers \footnote{We use conv1-1, conv1-2, conv2-1, conv2-2, conv3-1, conv3-2, conv3-3, conv4-1, conv4-2, conv4-3, conv5-1, conv5-2 and conv5-3 layers in the VGG network}. In the above equation, $\mathcal{C}$ refers to the set of conv layers in the network and $C_i$ is the feature vector obtained from the $i$th layer. The total cost function is the average similarity of all layers. The convolution filters in higher layers of the network provide abstract shape features while the details of the images such as edges and corners appear in the features obtained from the lower layers of the network. The features in higher levels have larger receptive fields, and can therefore cope with larger displacements, 
and help the optimization to converge in the first iterations when the initial estimates are far off. Similarly, the lower convolutional layers play a greater role in later iterations, to help the objects converge with more precision. In this way, the network provides a natural coarse-to-fine structure to the optimization.

Since our objective function is not differentiable we use COBYLA~\cite{powell1994direct}, a derivative free numerical optimization method, deployed in a Python optimization package. We found this procedure to work very well in practice. Figure~\ref{fig:optimization_convergence} shows the convergence of the method for example scenes.

\section{Coloring CAD models}
We use a medoid color of each object in the input image for scene optimization (Section \ref{sec:sceneOptimization}) and visualization. The process is as follows. First, we project the best aligned CAD model of an object onto its bounding box in the image. We then find the median value of each color channel separately, and take the closest color which appears within the mask. We also compute the medoid color for each wall of the room using a similar approach. We compute the mask of each wall through the room geometry, and exclude the bounding boxes from detected objects. This approach works well in practice for scene optimization and creates visually pleasant renderings without falling into the uncanny valley~\cite{seyama2007uncanny} (see results in Figure~\ref{fig:fullresults_1}).

%% file: result.tex
\section{Experimental Results}
\label{sec:result}
We evaluate our IM2CAD system both qualitatively and quantitatively on scene understanding benchmarks.

\subsection{Qualitative Evaluation}

We evaluated the proposed IM2CAD system with $100$ real world indoor images collected from ``Zillow Digs"~\cite{ZillowDigs}. These images are living room and bedroom shots as our training object categories are chair, table, sofa, bookshelf, bed, night table, chest, and window, i.e., typical bedroom and living room furniture. We cover a variety of room styles from traditional to modern with various furniture arrangement, complexity, and clutter that are representative of real world scenes. We also show example results on the SUN RGB-D dataset.

\begin{table}[t]
\centering
\begin{tabular}{lc}
\toprule
Method & Pixel Error(\%) \\
\midrule  
Lee et al.~\cite{dclee2009geometric}&24.70\\
Hedau et al.~\cite{Hedau09}& 21.20\\
Del Pero et al.~\cite{pero2012bayesian}& 16.30\\
Gupta et al.~\cite{gupta2010estimating}& 16.20\\
Zhao et al.~\cite{zhao2013scene}& 14.50\\
Schwing et al.~\cite{schwing2012efficient}& 13.59\\
Ramalingam et al.~\cite{ramalingam2013manhattan} & 13.34\\
Mallya et al.~\cite{mallya15}& 12.83\\
Dasgupta et al.~\cite{Dasgupta_2016_CVPR}& 9.73\\
Ren et al.~\cite{ren2016coarse}& 8.67\\
\midrule
IM2CAD & 10.15\\
\bottomrule
\end{tabular}
\vspace{-3 mm}
\caption {\small Room layout pixel misclassification error on Hedau~\protect\cite{Hedau09}.}
\label{tab:room_layout_quantitative}
  \vspace{-5 mm}
\end{table}

\begin{table}[t]
\centering
\begin{tabular}{lc}
\toprule
Method & Pixel Error(\%) \\
\midrule  
Hedau et al.~\cite{Hedau09}& 24.23\\
Mallya et al.~\cite{mallya15}& 16.71\\
Dasgupta et al.~\cite{Dasgupta_2016_CVPR}& 10.63\\
Ren et al.~\cite{ren2016coarse}& 9.31\\
\midrule
IM2CAD & 10.04\\
\bottomrule
\end{tabular}
\vspace{-3 mm}
\caption {\small Room layout pixel misclassification error on LSUN~\protect\cite{lsun}.}
\label{tab:room_layout_lsun}
  \vspace{-6 mm}
\end{table}

\begin{figure*}[t]
  \captionsetup[subfigure]{labelformat=empty,farskip=-10pt}
  \centering
  \subfloat[]{\includegraphics[width=.328\textwidth, height=.08\textheight]{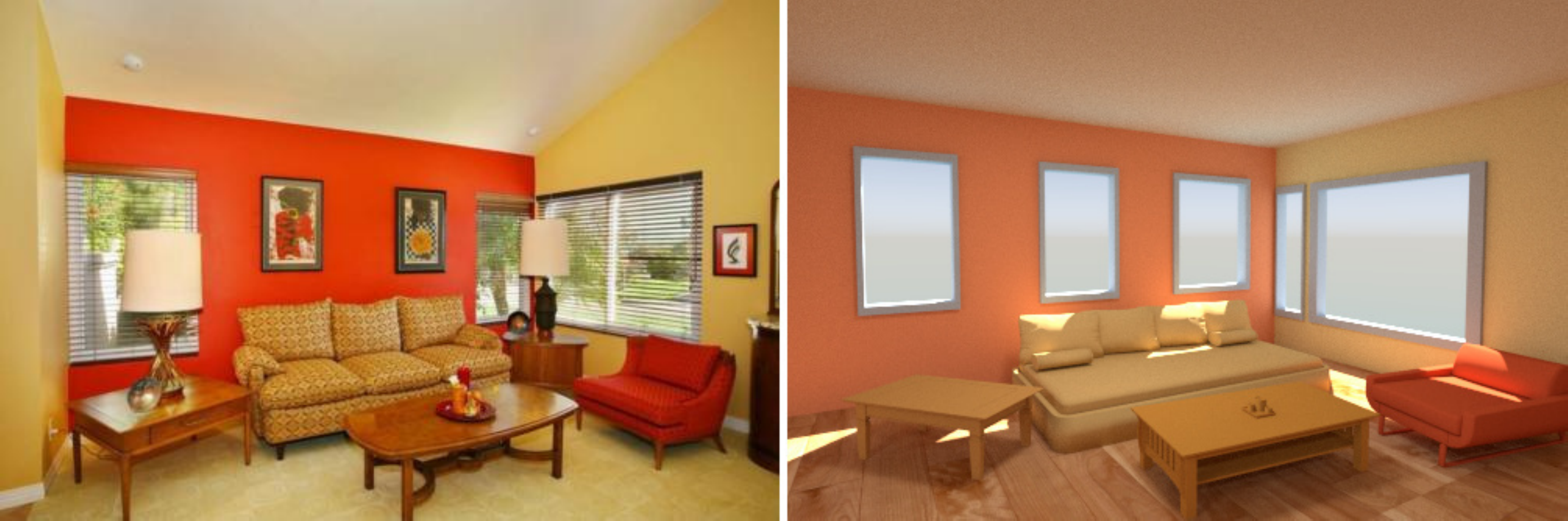}}~
  \subfloat[]{\includegraphics[width=.328\textwidth, height=.08\textheight]{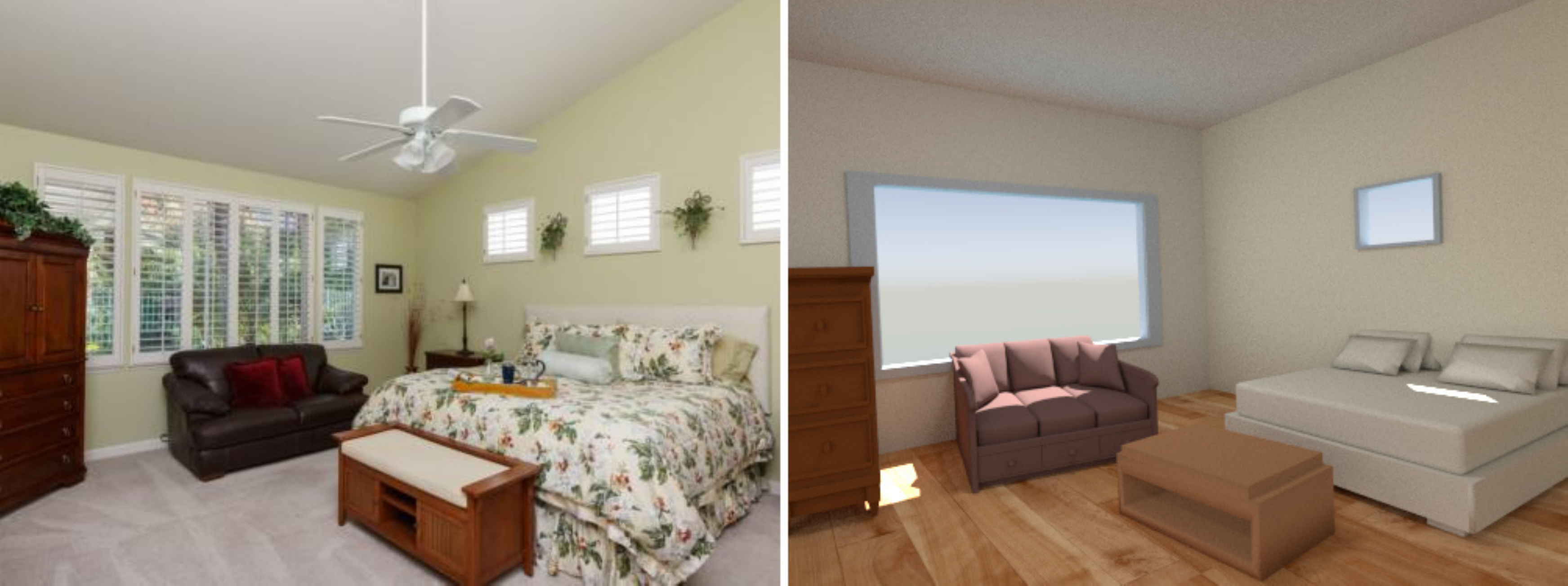}}~
  \subfloat[]{\includegraphics[width=.328\textwidth, height=.08\textheight]{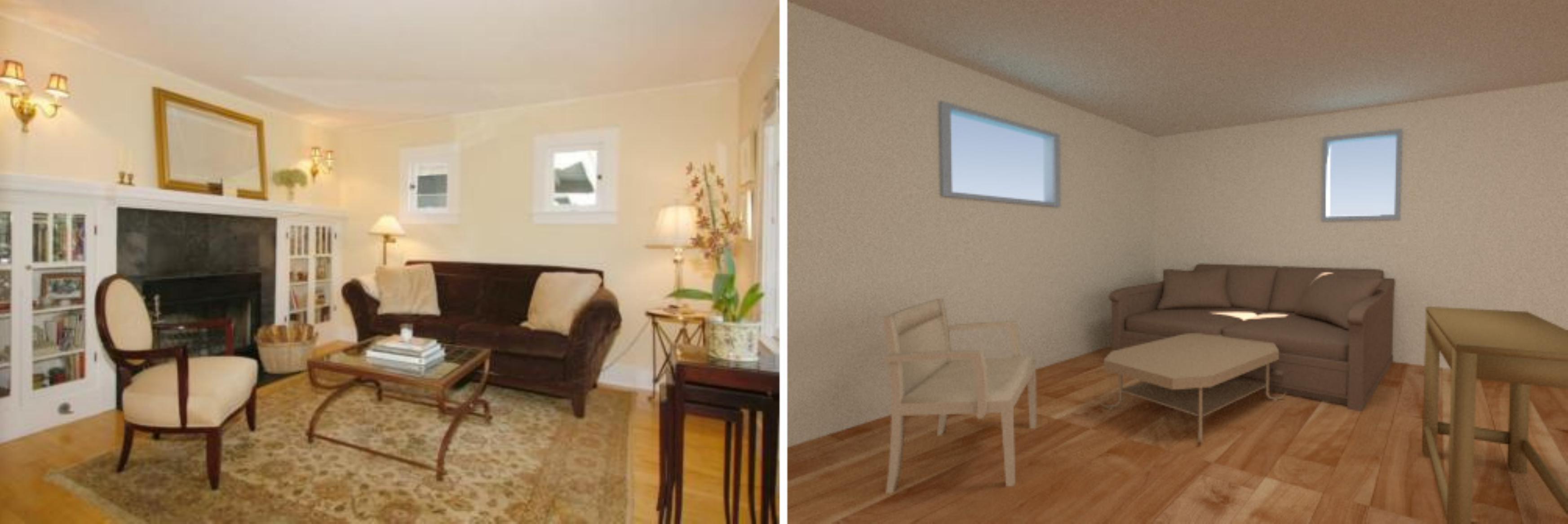}}\quad
  \subfloat[]{\includegraphics[width=.328\textwidth, height=.08\textheight]{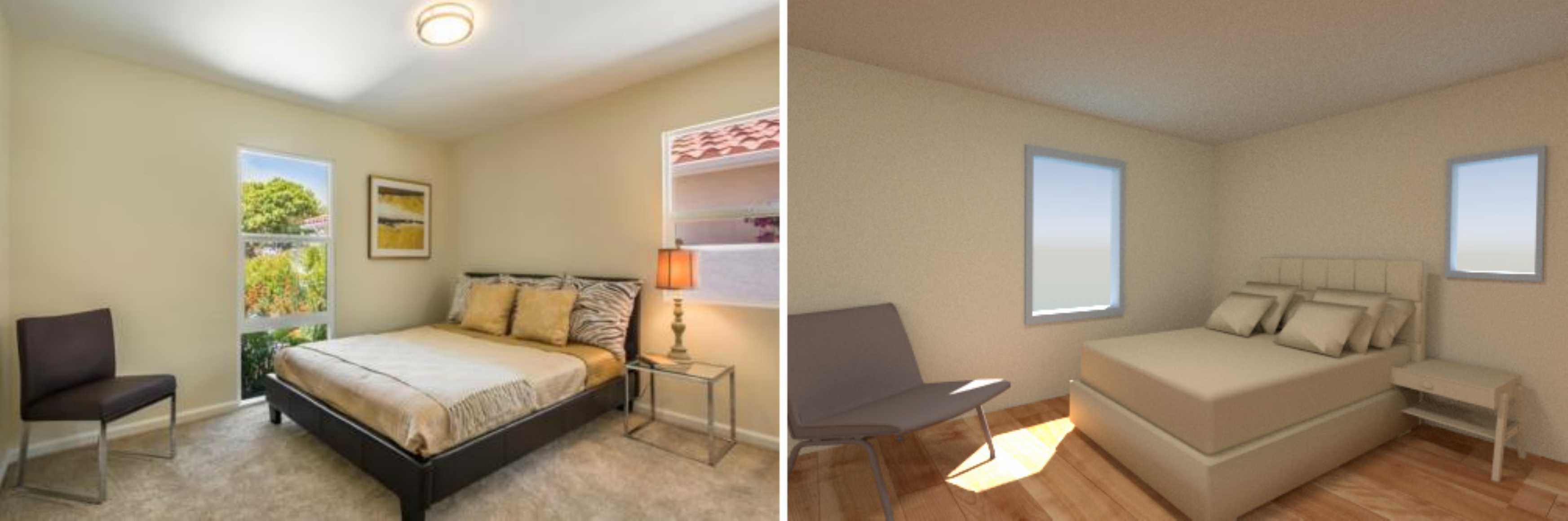}}~
  \subfloat[]{\includegraphics[width=.328\textwidth, height=.08\textheight]{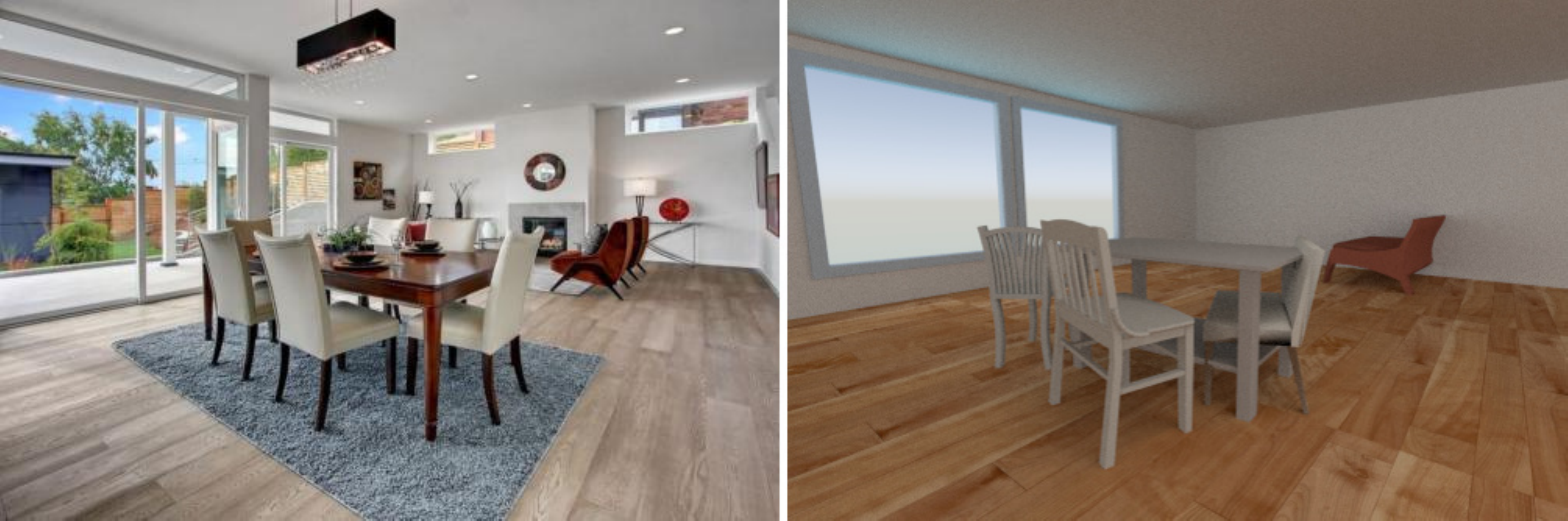}}~
  \subfloat[]{\includegraphics[width=.328\textwidth, height=.08\textheight]{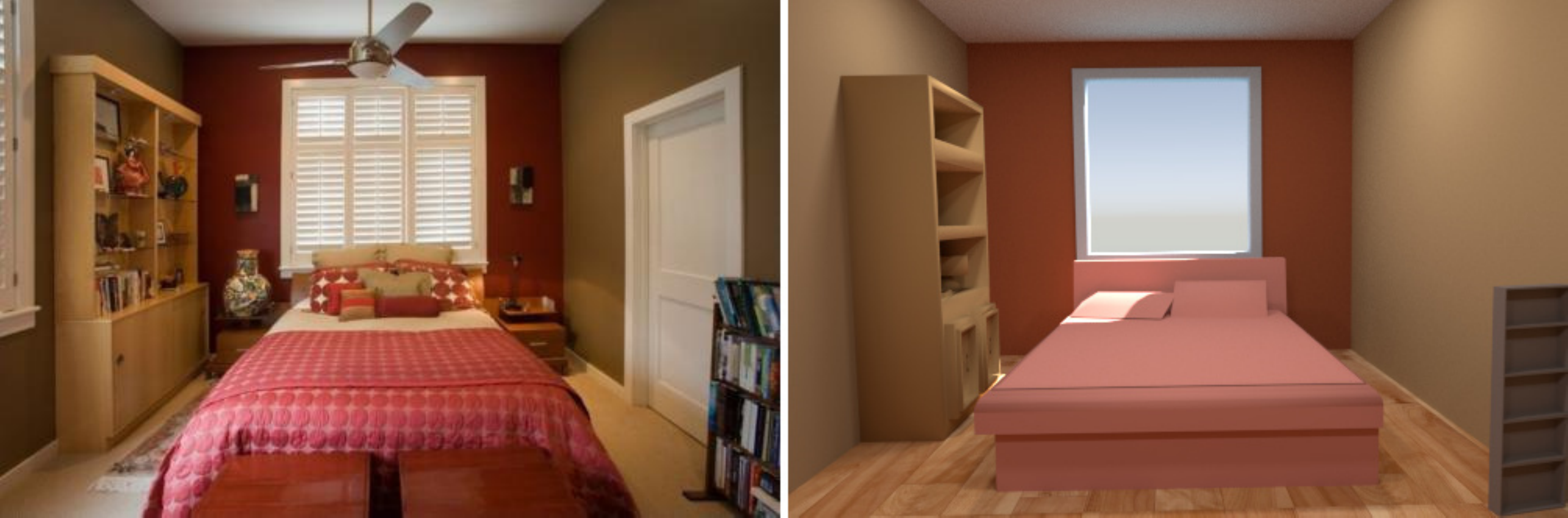}}\quad
  \subfloat[]{\includegraphics[width=.328\textwidth, height=.08\textheight]{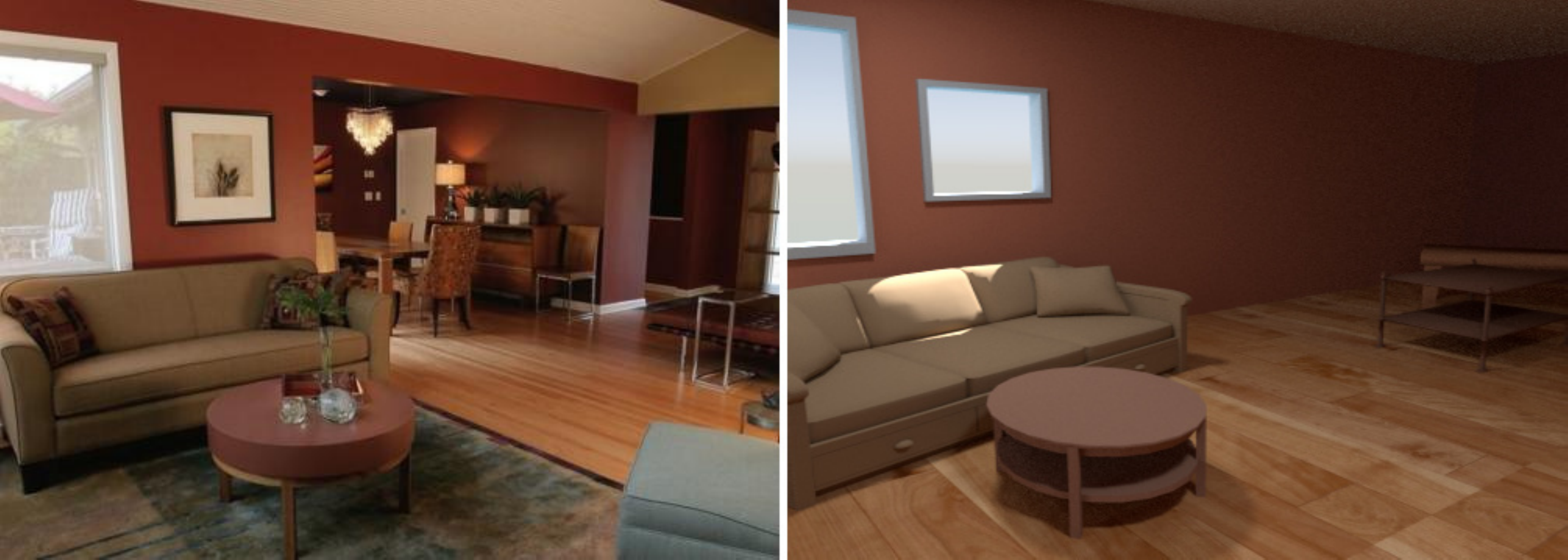}}~
  \subfloat[]{\includegraphics[width=.328\textwidth, height=.08\textheight]{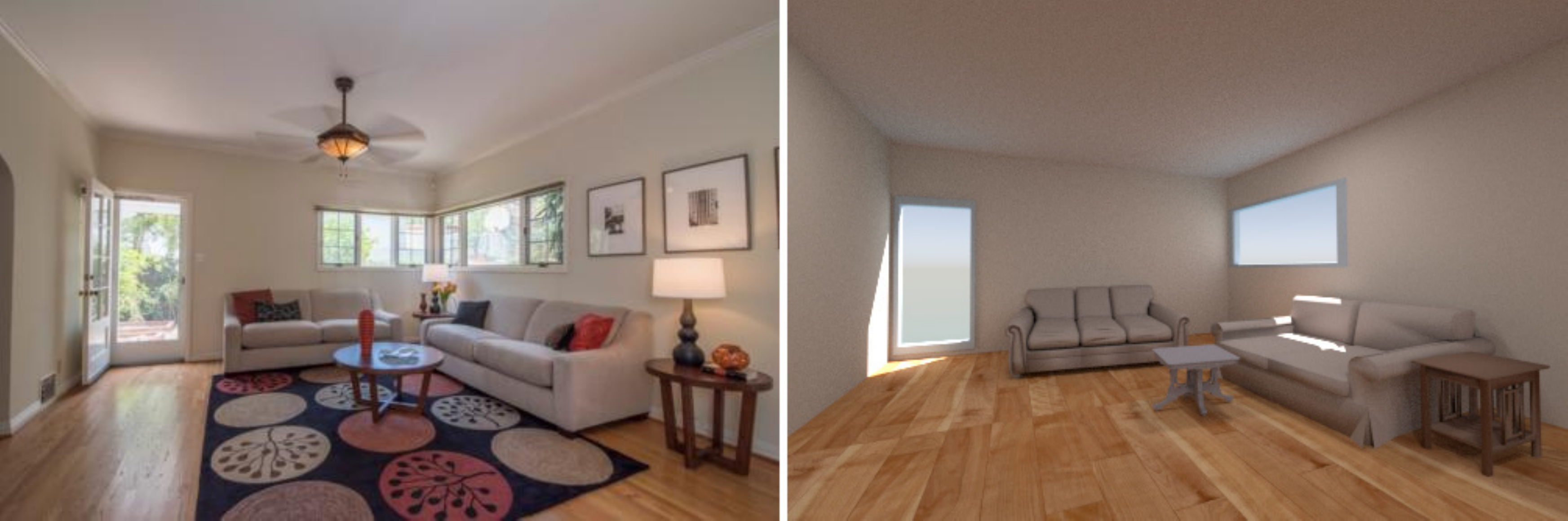}}~
  \subfloat[]{\includegraphics[width=.328\textwidth, height=.08\textheight]{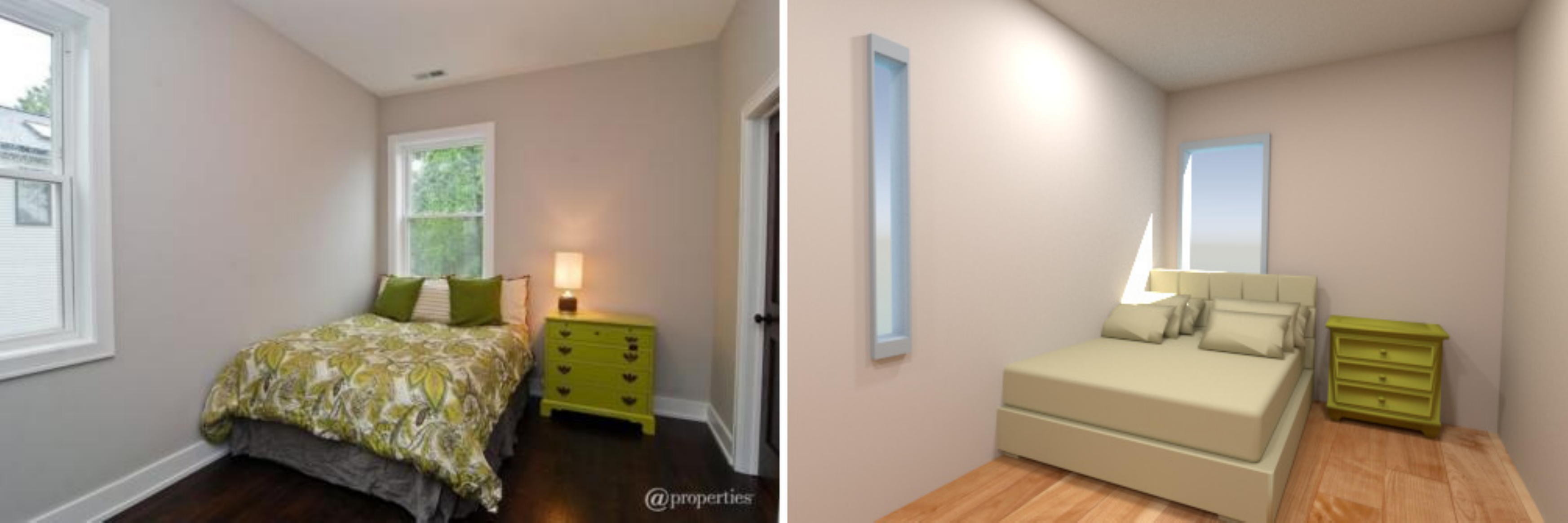}}\quad
  \subfloat[]{\includegraphics[width=.328\textwidth, height=.08\textheight]{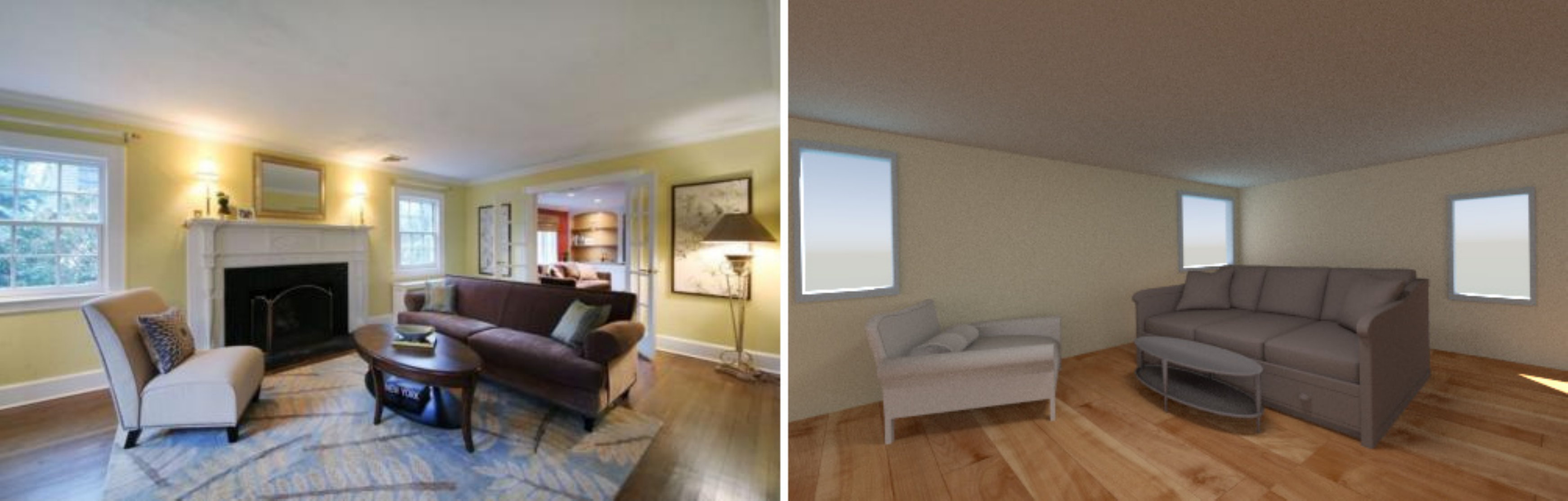}}~
  \subfloat[]{\includegraphics[width=.328\textwidth, height=.08\textheight]{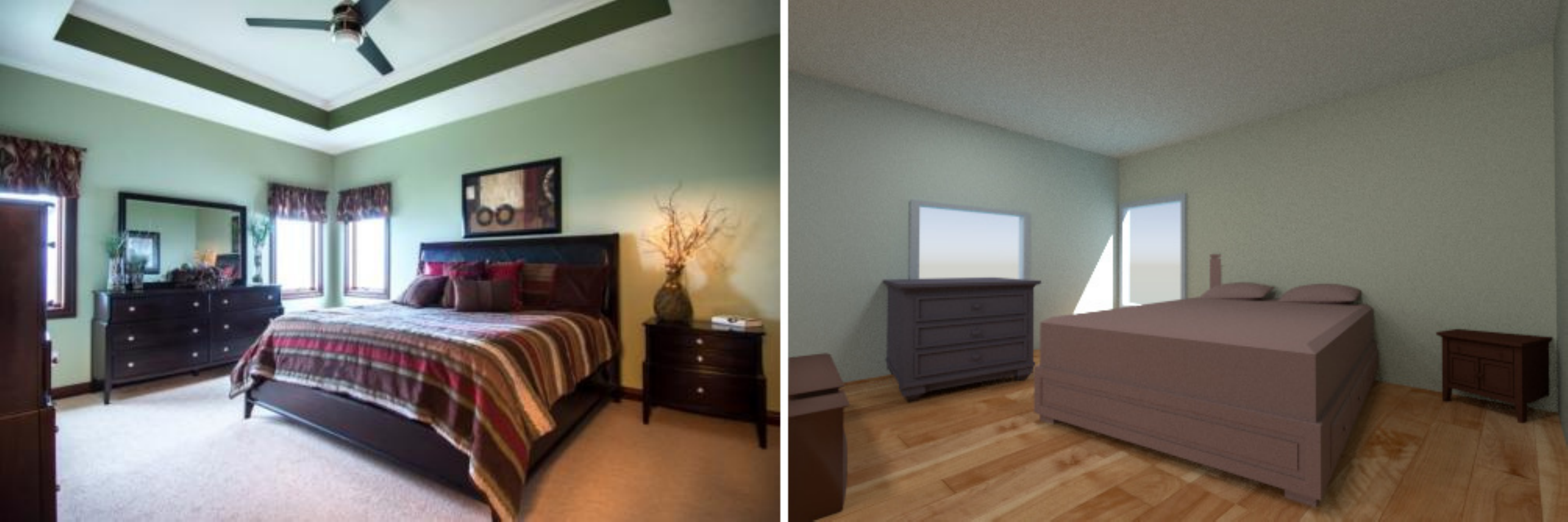}}~
  \subfloat[]{\includegraphics[width=.328\textwidth, height=.08\textheight]{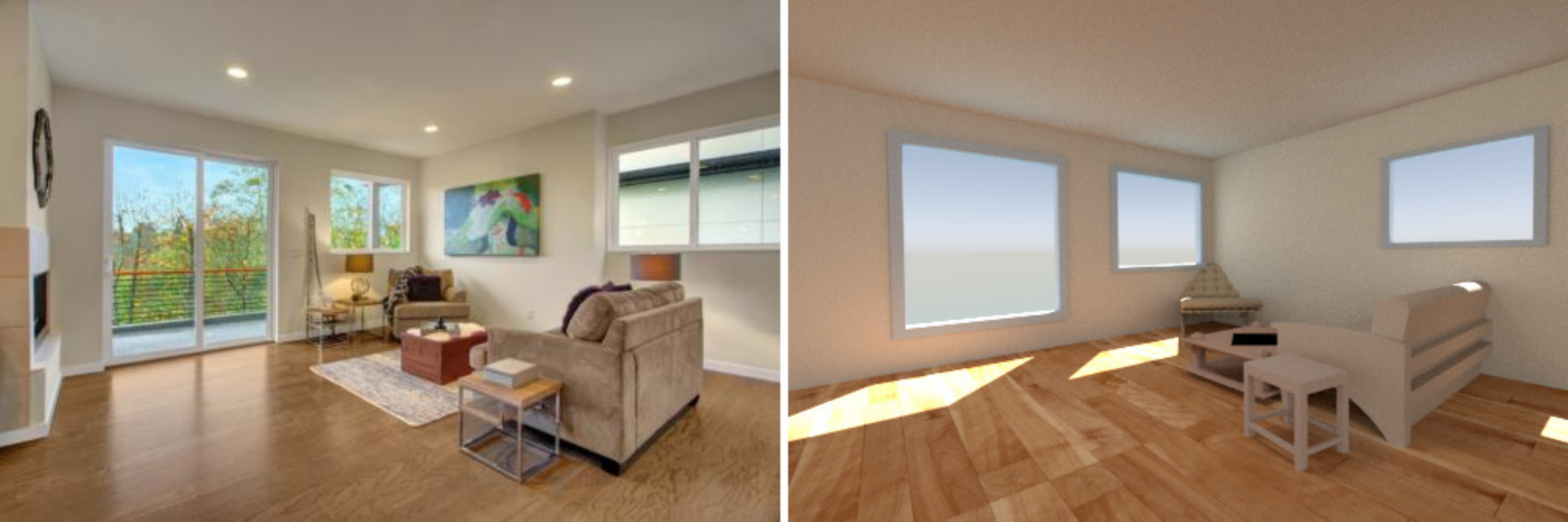}}\quad
  \subfloat[]{\includegraphics[width=.328\textwidth, height=.08\textheight]{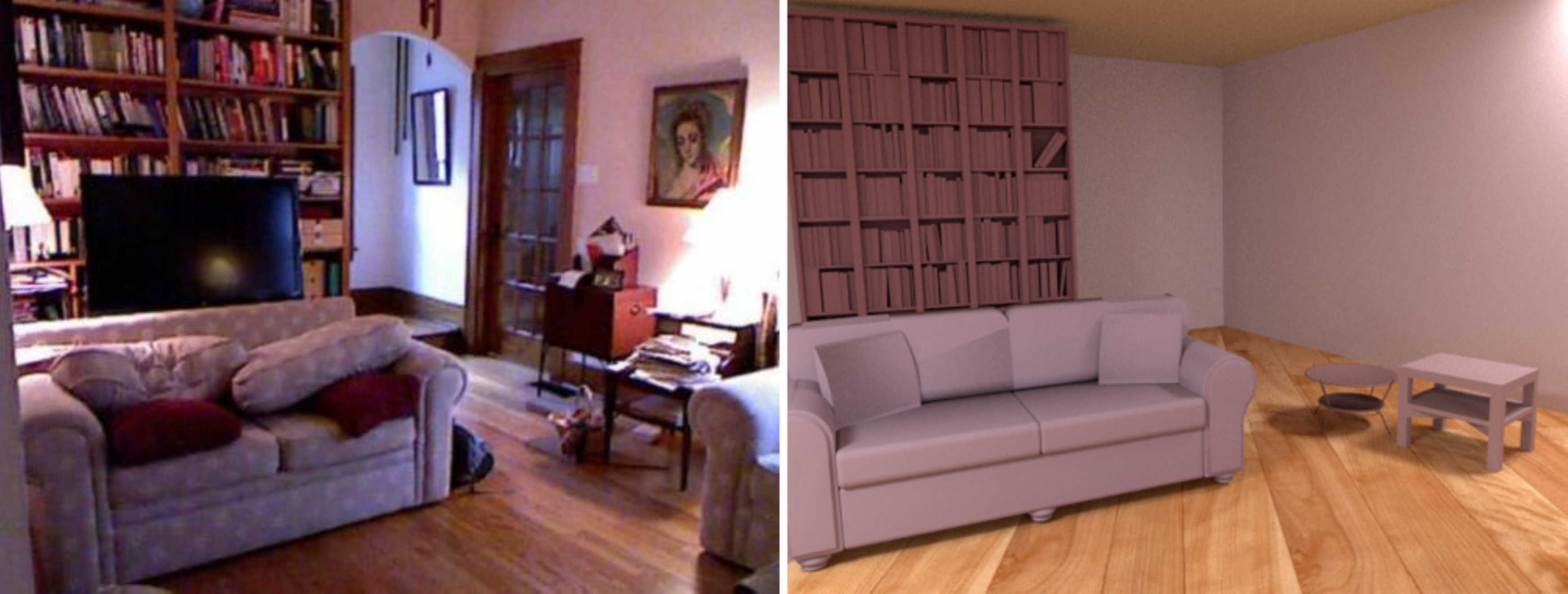}}~
  \subfloat[]{\includegraphics[width=.328\textwidth, height=.08\textheight]{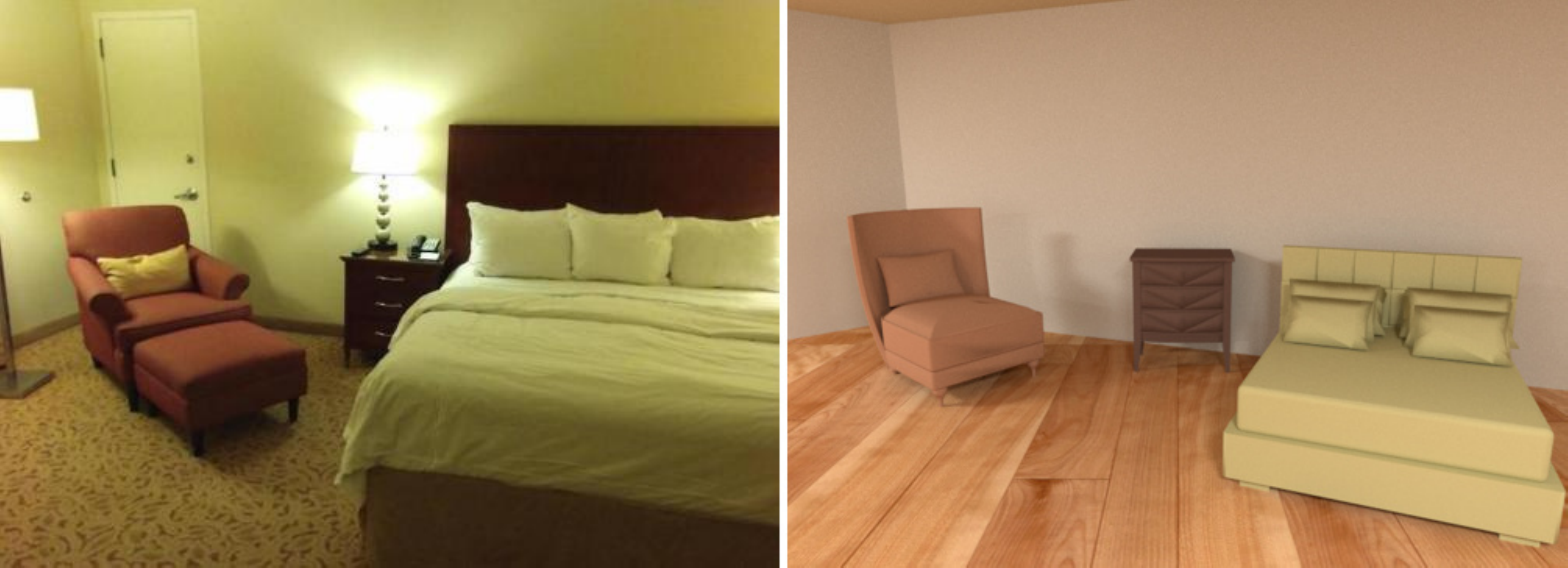}}~
  \subfloat[]{\includegraphics[width=.328\textwidth, height=.08\textheight]{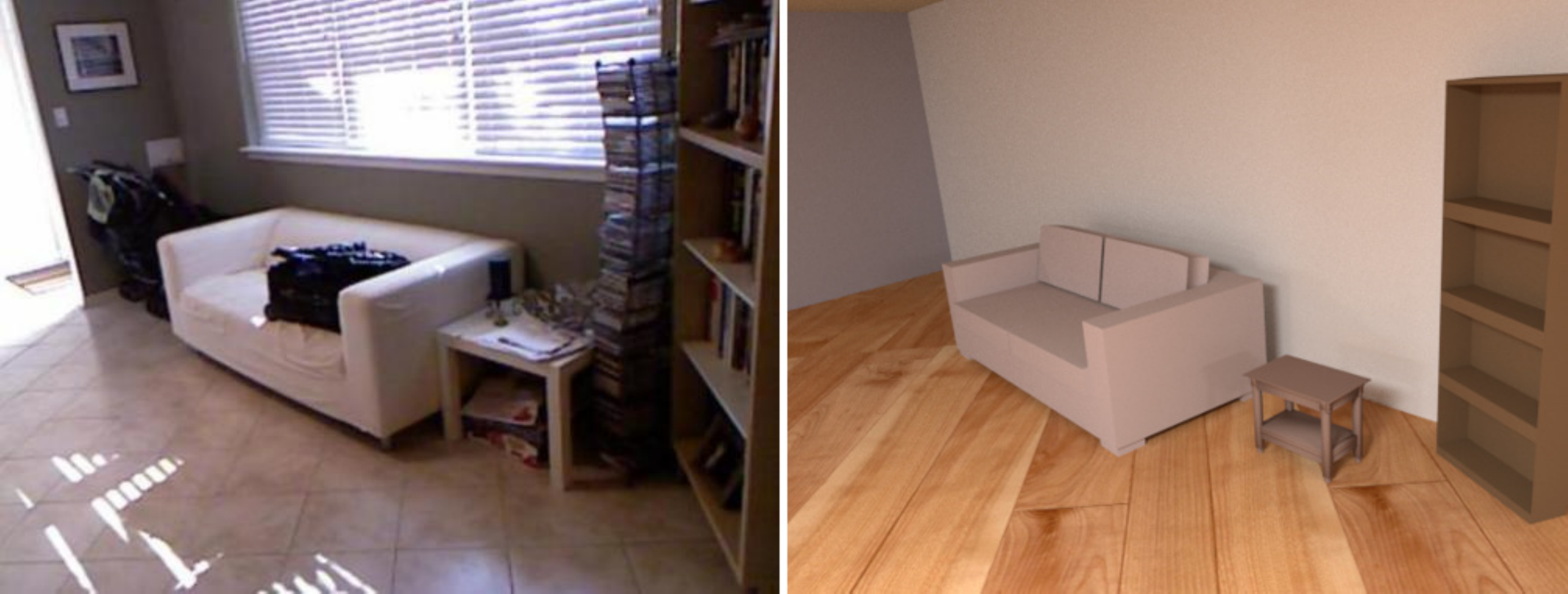}}\quad

  \vspace{-7 mm}
  \caption{The reconstruction results. In each example the left image is the real input image and the right image is the rendered 3D CAD model produced by IM2CAD. Last row shows example results on the SUN RGB-D dataset.}
\label{fig:fullresults_1}
  \vspace{-5 mm}
\end{figure*}

\begin{figure*}[t]
  \centering
  \captionsetup[subfigure]{farskip=-10pt}
  \centering
  \subfloat[]{\includegraphics[width=.33\textwidth]{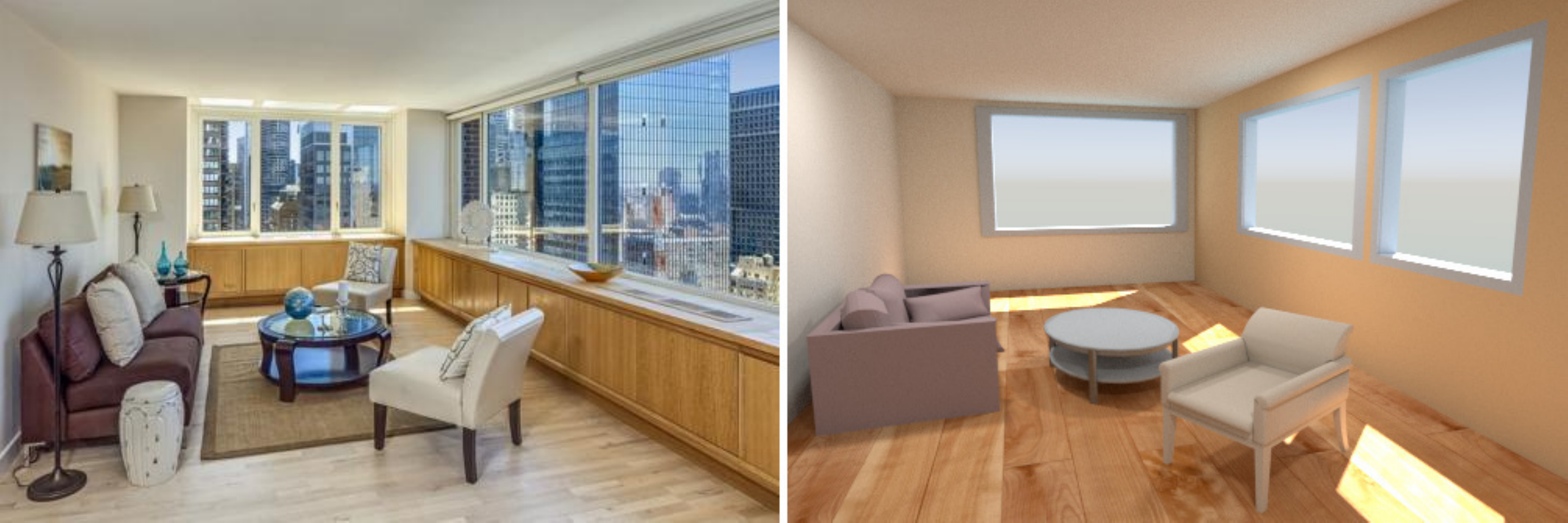}}~
  \subfloat[]{\includegraphics[width=.33\textwidth]{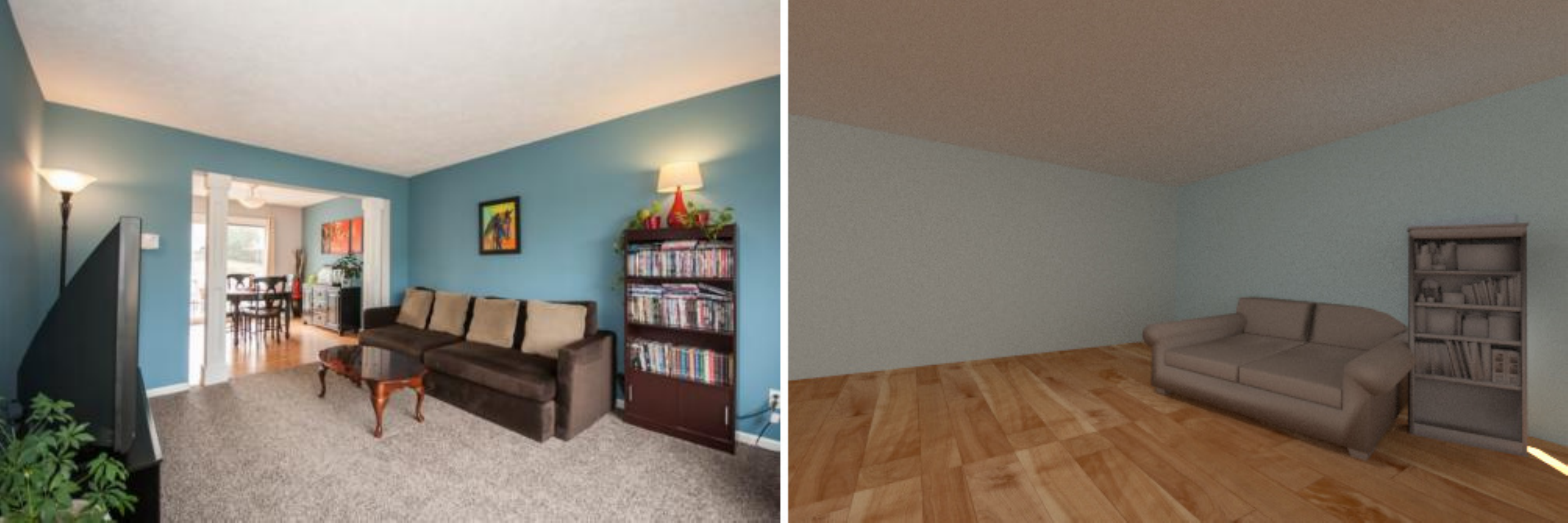}}~
  \subfloat[]{\includegraphics[width=.33\textwidth]{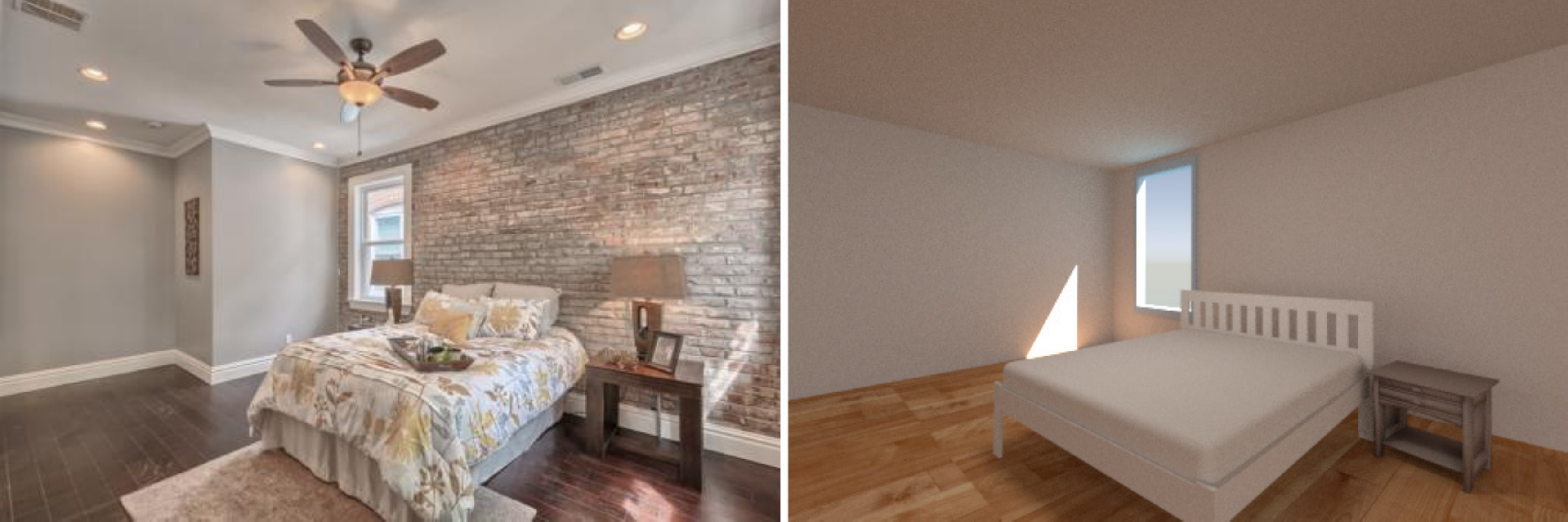}}
  \vspace{-4 mm}
  \caption{\small Failure cases: inaccurate chair pose (a); mis-detection of a chair (a) and table (b); non-cubic room shape (c).}
  \vspace{-6 mm}
\label{fig:failure}
\end{figure*}

Our IM2CAD approach consistently produces reasonable results on most of the test images. Figure~\ref{fig:fullresults_1} is representative of the top $30\%$ of our results, where most pieces of furniture are detected, represented using well-matched CAD models, and properly posed. Typical failure results are shown in Figure~\ref{fig:failure}.  Our failures are rarely catastrophic, and generally fall into the category of some furniture items being omitted or misplaced.

Object pose estimation can sometimes get stuck in local optimal. Notice that the foreground chair in Figure~\ref{fig:failure}(a) is in an incorrect pose while the chair legs are aligned almost perfectly to the image. The last two rows of Figure~\ref{fig:object_pose_estimation} demonstrate cases where the visual similarity fails to retrieve appropriate CAD models.  Heavily occluded objects impose additional challenges. Notice the missing chair and coffee table in Figure~\ref{fig:failure}(a) and (b).  If the room shape is not perfectly cubic (Figure~\ref{fig:failure}(c)),  room layout estimation can fail to recover the true room shape.  The windows can be confused with paintings as they have very similar visual features (see Figure~\ref{fig:fullresults_1}). Both windows and paintings typically appear as glassy and shiny rectangular shapes on a wall.

We use Caffe~\cite{jia2014caffe} to implement and train our deep networks. We learn the weights of our FCN network for the room geometry estimation using stochastic gradient descent with initial learning rate of 0.001 and weight decay of 5e-4. We train our network in 45 epochs where the learning rate decreases every 15 epochs. 
For the object detection we use same threshold for all object categories and only keep the detection boxes with scores higher that 0.5. 

Object detection and geometric feature extraction are processed on a Titan X GPU, while room layout sampling and object pose estimation are computed on CPU. For a typical input image of size $300 \times 500$, the computational time is approximately $0.15$ seconds for object detection, $0.3$ seconds for geometric feature extraction, $8$ seconds for room layout sampling and ranking, and $10$ seconds for object placement. Scene optimization is an iterative process where each iteration takes about $1$ second. We set the maximum number of iterations to be $250$. The overall CAD model creation process finishes within $5$ minutes. 

To produce final room renderings with global illumination, we use Blender Cycle Render Engine~\cite{blenderrenderer}, with fixed lighting consisting of distant sunlight from the top right point and five area lights on the ceiling. The final rendering process takes about $15$ minutes with global illumination.

\begin{table}
\centering
\begin{tabular}{lcc}
\toprule
Method &  SUN RGB-D & 3DGP\\
\midrule  
Hedau et al.~\cite{Hedau09}& 49.4 & 47.3\\
IM2CAD& 62.6 & 63.2\\
\bottomrule
\end{tabular}
\vspace{-3 mm}
\caption {3D room estimation results using voxel IoU on SUN RGB-D \protect~\cite{song2015sun} and 3DGP \protect~\cite{choi2015indoor} datasets (higher is better).}
\label{tab:3d_room_layout_quantitative}
  \vspace{-5 mm}
\end{table}

\subsection{2D Room Layout Estimation}
To evaluate the accuracy of room layout estimation, we compute the pixelwise difference between the predicted layout and the ground truth layout labels, averaged across all images as the evaluation metric. We evaluated on the test split of~\cite{Hedau09} dataset (we do not use their training split). Our FCN features (without 3D box estimation) achieve a $12.4\%$ pixel misclassification error compared to $28.9\%$ of~\cite{hoiem2007recovering} on the leading benchmark dataset~\cite{Hedau09} (see Figure~\ref{fig:geometric_feature}). 
When combined with a box-fitting step of~\cite{Hedau09,dclee2009geometric}, we achieve competitive result of $10.15\%$ error compared with~\cite{Dasgupta_2016_CVPR} and~\cite{ren2016coarse} as shown in Table~\ref{tab:room_layout_quantitative}. More specifically, we improve the reported result of~\cite{mallya15} by $2.7\%$, ~\cite{Dasgupta_2016_CVPR} by $3.1\%$, and~\cite{ren2016coarse} by $4.2\%$. As an ablation study to evaluate the effect of different room hypothesis estimation approaches, we tested our approach while being combined with either of~\cite{Hedau09} or~\cite{dclee2009geometric} and we obtain an error of $11.02\%$ and $11.13\%$, respectively.

\begin{table}
\centering
\begin{tabular}{l<{\hspace{-14pt}}c<{\hspace{-5pt}}c<{\hspace{-5pt}}c<{\hspace{-5pt}}c<{\hspace{-4pt}}}
\toprule
 & \multicolumn{2}{c}{SUN RGB-D}  & \multicolumn{2}{c}{3DGP} \\
Method & voxel IoU & mAP & voxel IoU & mAP\\
\midrule  
3DGP~\cite{choi2015indoor}& 38.7&42.1& 38.4 & 59.7\\
IM2CAD (w/o optim.)& 46.1&74.7& 53.5 & 86.6\\
IM2CAD (w/ optim.)& 49.0&75.6& 53.8 & 86.6\\
\bottomrule
\end{tabular}
\vspace{-3 mm}
\caption {3D scene free space prediction (voxel IoU) and object localization (mAP) results on SUN RGB-D \protect~\cite{song2015sun} and 3DGP \protect~\cite{choi2015indoor} datasets (higher is better).}
\label{tab:3d_scene_quantitative}
  \vspace{-5 mm}
\end{table}

We also evaluated performance on the task of room layout pixel misclassification using the LSUN dataset~\cite{lsun}. As summarized in Table~\ref{tab:room_layout_lsun}, IM2CAD outperforms previous approaches ~\cite{Hedau09,mallya15} significantly as well as ~\cite{Dasgupta_2016_CVPR} and obtains competitive results with recent approach of~\cite{ren2016coarse}.

\subsection{3D Room Estimation and Scene Understanding}
Our IM2CAD system is also applicable for 2D and 3D scene understanding as well as room layout estimation. For evaluating our performance in scene understanding tasks, we use the SUN RGB-D dataset~\cite{song2015sun}. 
This dataset contains images captured from different view points, some of the images have low field of view and a considerable number of them are captured from highly cluttered scenes. Note that, although the SUN RGB-D dataset contains the depth data for all the images, we do not use the depth information at either train or test time, but estimate the 3D room geometry as well as object layout using only single 2D images. We use the test split for the bedroom and living room scene categories with a total of 484 images.

\noindent{\bf 3D Room Layout Estimation}
~ ~3D room layout estimation enables precise reasoning about free space versus spaces occupied by objects. In the absence of depth data, this task is challenging as it requires reasoning about room geometry from 2D images. Our 3D room layout estimation is evaluated by computing the intersection over union (IoU) between the predicted and the ground truth free spaces. Following~\cite{song2015sun}, we assume empty rooms without objects and define a voxel grid of $0.1\times0.1\times0.1$ meter. The effective voxels are the ones that are located within $0.5$ and $5.5$ meters from the camera and are inside the field of view. We check whether each voxel is inside the 3D room polygon and compute the intersection and union computed by counting the 3D voxels. 
Table~\ref{tab:3d_room_layout_quantitative} summarizes our obtained results. Our method outperforms ~\cite{Hedau09} by $13.2\%$.

\noindent{\bf Scene Understanding}
~ ~The task of scene understanding integrates recognition and localization of all the objects as well as estimating the room structure. Compared to the task of 3D room estimation, this is a more challenging task as it requires detecting non-free spaces occupied by the objects. We compute the distance between the projection of the box centroid on the ground plane for all pairs of predicted and ground truth objects with the same label. We sort the distances in ascending order for each available pair and choose the pair with the shortest distance while the two boxes are marked as unavailable. We compute the precision and recall by varying the distance threshold and use the mean average precision as object localization metric.

Free space prediction is evaluated in a similar manner to the 3D room layout. The visible 3D voxels for the free space inside the room polygon but outside any object bounding box is computed and then the IoU between the free space prediction and the ground truth is computed. Table~\ref{tab:3d_scene_quantitative} shows the results of free space prediction and object localization on SUN RGB-D dataset. We compare the performance of our approach for scene understanding with~\cite{choi2015indoor}. IM2CAD obtains superior results compared with ~\cite{choi2015indoor} in both metrics i.e., $33.5\%$ boost in the mean AP and $11.7\%$ in scene free space prediction. We compare our results before and after applying scene optimization (Section ~\ref{sec:sceneOptimization}). Our scene optimization approach results in improved accuracy for the task of scene understanding.

We also report IM2CAD performance on the dataset presented in~\cite{choi2015indoor} which we call 3DGP. We use 372 images from the test split of living room, bedroom and dining room categories. However, we do not train our model on the 3DGP training set. To estimate the ground truth camera parameters, we compute the pseudo ground truth vanishing points by using the annotated ground truth edges corresponding to the three vanishing points following the experimental setting of~\cite{choi2015indoor} for 3D scene evaluation. We evaluate on the three tasks of 3D room layout, whole scene free space prediction, and object localization. These results are summarized in Tables~\ref{tab:3d_room_layout_quantitative} and~\ref{tab:3d_scene_quantitative}. For the task of 3D room estimation, IM2CAD significantly outperforms~\cite{choi2015indoor} by $15.9\%$. In the free space prediction task, IM2CAD obtains significantly better results than 3DGP in both voxel IoU and mean AP criteria.

%% file: conclusion.tex
\section{Conclusion}
\label{sec:conclusion}

This paper presents a fully automatic system that reconstructs a 3D CAD model of an indoor scene from a single photograph, by utilizing a large database of 3D furniture models. It estimates room geometry, and detects and aligns objects in the image with accurate 3D poses.
We introduce novel approaches for room modeling and scene optimization, that are keys to the success of our system. We evaluate on a wide range of living room and bedroom photographs with a variety of home styles. The results demonstrate the effectiveness of our approach in creating 3D CAD models that faithfully resemble the real scenes. With the abundance of indoor photos available online, our system is applicable to produce a large database of indoor scene models. Our approach obtains significant improvement on the 2D room layout estimation and 3D scene understanding benchmarks. 

Our system does have limitations suggesting a number of areas for future work. We assume the room geometry in the image can be modeled with a cube. Working with complicated room geometry is an area of future improvement. Understandably, heavily occluded objects impose challenges.  We assume objects are always on the ground plane (e.g., chairs and beds) or attached to walls (windows), posing a lamp on a table would require extension of our work. Incorporating more object types would lead to more general scenes and room types (e.g. kitchens and bathrooms).